\documentclass{article}

\PassOptionsToPackage{numbers, compress}{natbib}

\usepackage[preprint]{neurips_2021}

\usepackage{amsmath,amsfonts,bm}

\def\eqref#1{equation~\ref{#1}}

\def\1{\bm{1}}

\DeclareMathAlphabet{\mathsfit}{\encodingdefault}{\sfdefault}{m}{sl}
\SetMathAlphabet{\mathsfit}{bold}{\encodingdefault}{\sfdefault}{bx}{n}

\newcommand{\E}{\mathbb{E}}

\newcommand{\R}{\mathbb{R}}

\DeclareMathOperator*{\argmax}{arg\,max}
\DeclareMathOperator*{\argmin}{arg\,min}

\usepackage[utf8]{inputenc} 
\usepackage[T1]{fontenc}    
\usepackage{hyperref}       
\usepackage{url}            
\usepackage{booktabs}       
\usepackage{amsfonts}       
\usepackage{nicefrac}       
\usepackage{microtype}      
\usepackage{xcolor}         
\usepackage{graphicx}
\usepackage{mathtools}
\usepackage[english]{babel}

\usepackage{setspace}
\usepackage{wrapfig}
\usepackage{algorithm}
\usepackage{algorithmic}
\usepackage{comment}
\usepackage{csquotes}
\usepackage{subcaption}

\renewcommand{\eqref}[1]{(\ref{#1})}  

\title{Towards the Memorization Effect of Neural Networks in Adversarial Training}

\author{%
  Han Xu, Xiaorui Liu, Wentao Wang, Anil K. Jain. Jiliang Tang\\ 
  Department of Computer Science and Engineering\\
  Michigan State University\\
  \And Wenbiao Ding, Zhongqin Wu, Zitao Liu\\
  TAL Education Group\\
  
}

\begin{document}

\maketitle

\begin{abstract}
Recent studies suggest that ``memorization'' is one important factor for overparameterized deep neural networks (DNNs) to achieve optimal performance. Specifically, the perfectly fitted DNNs can memorize the labels of many atypical samples, generalize their memorization to correctly classify test atypical samples and enjoy better test performance. While, DNNs which are optimized via adversarial training algorithms can also achieve perfect training performance by memorizing the labels of atypical samples, as well as the adversarially perturbed atypical samples. However, adversarially trained models always suffer from poor generalization, with both relatively low clean accuracy and robustness on the test set. In this work, we study the effect of memorization in adversarial trained DNNs and disclose two important findings: \textbf{(a)} Memorizing atypical samples is only effective to improve DNN's accuracy on clean atypical samples, but hardly improve their adversarial robustness and \textbf{(b)} Memorizing certain atypical samples will even hurt the DNN's performance on typical samples. Based on these two findings, we propose \textit{Benign Adversarial Training (BAT)} which can facilitate adversarial training to avoid fitting ``harmful'' atypical samples and fit as more ``benign'' atypical samples as possible. In our experiments, we validate the effectiveness of BAT, and show it can achieve better clean accuracy vs. robustness trade-off than baseline methods, in benchmark datasets such as CIFAR100 and Tiny~ImageNet.

\end{abstract}

\vspace{-0.3cm}
\section{Introduction}
It is evident from recent studies that the memorization effect (or benign overfitting)~\cite{feldman2020does, feldman2020neural, bartlett2020benign, chatterji2020finite, muthukumar2020harmless} can be one necessary factor for overparametrized deep neural networks (DNNs) to achieve the close-to-optimal generalization error. From the empirical perspective, the works~\cite{feldman2020does, feldman2020neural} suggest that modern benchmark datasets, such as CIFAR10, CIFAR100~\cite{krizhevsky2009learning} and  ImageNet~\cite{krizhevsky2012imagenet}, always have very diverse data distributions, especially containing a large fraction of ``atypical'' samples. These atypical samples are both visually and statistically very different from the other samples in their labeled class. For example, the images in the class ``bird'' may have a variety of sub-populations or species, with many samples in the main sub-population and other (atypical) samples in less-frequent and distinct sub-populations. Since these atypical samples are deviated from the main sub-population, DNNs can only fit these atypical samples by ``memorizing'' their labels. While, memorizing or fitting these atypical samples will not hurt the DNNs' performance, but can help DNNs to correctly classify the atypical samples appearing in the test set and hence improve the test accuracy.




Similar to the classification models which are trained via empirical risk minimization (ERM) algorithms, adversarial training methods
~\cite{madry2017towards, kurakin2016adversarial} are also devised to fit the whole training dataset. Specifically, adversarial training minimizes the model's error against adversarial perturbations~\cite{goodfellow2014explaining, szegedy2013intriguing} by fitting the model on manually generated adversarial examples of all training data. Although adversarial training can fit and memorize all training data as well as their adversarially perturbed counterparts, they always suffer from both poor clean accuracy and adversarial accuracy (or robustness)\footnotemark on the test set~\cite{tsipras2018robustness, schmidt2018adversarially}. Recent study~\cite{rice2020overfitting} indicates that, during the adversarial training process, one model's test adversarial accuracy will even keep decreasing as it hits higher training adversarial accuracy (during the fine-tuning epochs). Thus, it is natural to ask a question: \textit{What is the effect of memorization in adversarial training? In particular, can memorizing (the labels of) the atypical samples and their adversarial counterparts benefit the model's accuracy and robustness?}

\footnotetext{Clean Accuracy: Model's accuracy on unperturbed samples. Adversarial accuracy: Model's accuracy on adversarially perturbed samples. Without loss of generality, this paper discusses the adversarial accuracy under $l_\infty$-8/255 PGD attack~\cite{madry2017towards}.} 

To answer this question, we first conduct preliminary studies to check whether memorizing atypical samples in adversarial training can benefit the DNNs' test performance, especially on those test atypical samples. In Section~\ref{sec:pre1}, we implement PGD adversarial training~\cite{madry2017towards} on CIFAR100 under ResNet18 and WideResNet models~\cite{he2016deep} and fine-tune them until achieving the optimal training performance.
From the results in Section~\ref{sec:pre1}, we observe that {\it the memorization in adversarial training can only benefit the clean accuracy of test atypical samples.} With the DNNs gradually fit/memorize more atypical samples, they can finally achieve fair clean accuracy close to $\sim40\%$ on test atypical set.
However, the adversarial accuracy on test atypical set is constantly low ($\sim10\%$) during the whole training process, even though the models can fit almost all atypical (adversarial) training samples. Based on the theoretical study~\cite{schmidt2018adversarially}, the adversarial robustness is hard to generalize especially when the training data size is limited. Since each single atypical sample is distinct from the main sub-population and has rare frequency to appear in the training set, the data complexity for each specific atypical sample is also very low. Thus, its adversarial robustness can be extremely hard to generalize. Remarkably, the entire atypical set covers a non-ignorable fraction in complex datasets such as CIFAR100. Completely failing on atypical samples could be one important reason that contributes to the overall poor robustness generalization of DNNs.

Furthermore, we find that {\it in adversarial training, fitting atypical samples will even hurt DNNs' performance of those ``typical'' samples (the samples in the main sub-population)}. In the Section~\ref{sec:pre2}, we again implement PGD adversarial training~\cite{madry2017towards} on CIFAR100 for several trails, which are trained with different amount of atypical samples existed. 
Based on the results from Section~\ref{sec:pre2}, an adversarially trained model on the training set without any atypical samples has 92\% clean accuracy and 52\% adversarial accuracy on the test typical set. While, the model trained with 100\% atypical samples included only has 85\%/44\% clean/adversarial accuracy respectively. In other words, atypical samples act more like \textit{``poisoning data''}~\cite{biggio2012poisoning} to deteriorate model performance on typical samples. In this paper, we demonstrate that the atypical samples which ``poison'' the models are the ones which deviated from the distribution of their own classes, but close to the distribution of other classes. Some examples are shown in Fig.~\ref{fig:atypical_samples}. For instance, an atypical ``plate'' in CIFAR100 has strong visual similarity to ``apples''. If DNNs memorize the features of this ``plate'' and predict any other images with similar features to be ``plate'', the DNNs cannot distinguish ``apples'' and ``plates''.

Motivated by our findings, we propose a novel algorithm called \textit{Benign Adversarial Training (BAT)}, which can prevent the negative influence brought by memorizing ``poisoning'' atypical samples, meanwhile preserve the model's ability to memorize those ``benign / useful'' atypical samples. It is worth mentioning that, by fitting those ``benign'' atypical samples, the BAT method can achieve good clean accuracy on the atypical set. Compared with PGD adversarial training~\cite{madry2017towards} when it achieves the highest adversarial robustness, BAT has higher clean accuracy and comparable adversarial accuracy. Compared to other popular variants of adversarial training such as~\cite{zhang2019theoretically, wang2019improving, zhang2020geometry}, BAT is the only one obtaining both better (or comparable) clean and adversarial accuracy than~\cite{madry2017towards}, on complex datasets such as CIFAR100 and Tiny~Imagenet~\cite{le2015tiny}.

\vspace{-0.3cm}
\section{Definition and Notation}
\vspace{-0.1cm}
\subsection{Atypical Samples and Memorization}\label{sec:def_atypical}
In this section, we start by introducing necessary concepts and definitions about the memorization effects. As well known, overparameterized DNNs have tremendous capacity to make them easy to perfectly fit the training dataset~\cite{zhang2016understanding}. In practice, on common benchmark classification tasks, such as CIFAR10~\cite{he2016deep}, CIFAR100 and ImageNet~\cite{krizhevsky2012imagenet}, we also tune the DNNs to achieve very high training accuracy even close to 100\%, and enjoy good test performance. While, this property of DNNs in practice cannot be well explained by standard theories about model generalization~\cite{evgeniou2000regularization, bartlett2002rademacher} from the regularization perspective. At a high level, the standard theories underline the importance of regularization on the model complexity, to make DNNs to avoid ``overfitting'' or ``memorizing'' the outliers and nonuseful samples in the training data. 

Fortunately, recent works~\cite{feldman2020does, feldman2020neural,bartlett2020benign, muthukumar2020harmless, chatterji2020finite} make significant progress to close this gap from both theoretical and empirical perspectives. They suggest that the memorization effect is one necessary property for DNNs to achieve optimal generalization performance. In detail, the empirical studies~\cite{feldman2020does, feldman2020neural} point out that the common benchmark datasets, such as CIFAR10, CIFAR100, ImageNet, contain a large portion of atypical samples (or namely, rare samples, sub-populations, etc.). These atypical samples look very different from the other samples in the main distribution of its labeled class (see Appendix~\ref{app:atypical}), and are statistically indistinguished from outliers or mislabeled samples. Because these atypical samples are deviated from the main distribution, DNNs can only fit these samples by memorizing their labels. Moreover, without memorizing these atypical samples during training, the DNNs can totally fail to predict the atypical samples appearing in the test set~\cite{feldman2020neural}.

\noindent\textbf{Identify Atypical Samples.} To identify such atypical samples in common datasets in practice, the work~\cite{feldman2020neural} proposes to examine which training samples can only be fitted by memorization, and measure each training sample's \textit{``memorization value''}. Formally, for a training algorithm $\mathcal{A}$ (i.e., ERM), the memorization value ``$\text{mem}(\mathcal{A}, \mathcal{D}, x_i)$'' of a training sample $(x_i,y_i)\in \mathcal{D}$ in training set $\mathcal{D}$ is defined as:
\begin{align}\label{eq:mem}
    \text{mem}(\mathcal{A}, \mathcal{D}, x_i) = \underset{F\leftarrow \mathcal{A}(\mathcal{D})}{\text{Pr.}} (F(x_i) = y_i) - \underset{F\leftarrow \mathcal{A}(\mathcal{D}\backslash x_i)}{\text{Pr.}} (F(x_i) = y_i),
\end{align}
which calculates the difference between the model $F$'s accuracy on $x_i$ with and without $x_i$ removed from the training set $\mathcal{D}$ of algorithm $\mathcal{A}$. 
Note that for each sample $x_i$, if its memorization value is high, it means that removing $x_i$ from training data will cause the model with a high possibility to wrongly classify itself, so $x_i$ is very likely to be fitted only by memorization and be atypical. From the work~\cite{feldman2020neural}, these atypical samples have non-ignorable portion in common datasets. For example, there are around $40\%$ training samples in CIFAR100 having a large memorization value $>0.15$. 

The similar strategy can also facilitate to find atypical samples in the test set, which are the samples that are strongly influenced by atypical training samples.
In detail, by removing an atypical training sample $(x_i, y_i)$, we calculate its \textit{``influence value''} on each test sample $(x_j', y_j')\in\mathcal{D}'$ in test set $\mathcal{D}'$:
\begin{align}\label{eq:infl}
    \text{infl}(\mathcal{A}, \mathcal{D}, x_i, x'_j) = \underset{F\leftarrow \mathcal{A}(\mathcal{D})}{\text{Pr.}} (F(x_j') = y_j') - \underset{F\leftarrow \mathcal{A}(\mathcal{D}\backslash x_i)}{\text{Pr.}} (F(x_j') = y_j').
\end{align} 
If the sample pair $(x_i,x'_j)$ has a high influence value, removing the atypical sample $x_i$ will drastically decrease the model's accuracy on $x'_j$. It suggests that the model's prediction on $x'_j$ is mainly based on the memorization of $x_i$, thus, $x'_j$ is the corresponding atypical sample of $x_i$. The sample pair $x_i$ and $x'_j$ is called a \textit{high-influence pair}, if they have a high influence value and belong to the same class. In practice, the high-influence pairs of images are typically visually similar and have similar semantic features. The memorization benefits the model's performance especially on these test atypical samples, and hence boosts the overall test accuracy. 

\vspace{-0.2cm}
\subsection{Adversarial Training}
Similar to classification models trained via empirical risk minimization (ERM) algorithms, adversarial training methods
~\cite{madry2017towards, kurakin2016adversarial} are also devised to fit the whole dataset by training the model on manually generated adversarial examples. Formally, they are optimized to have the minimum adversarial loss:
\begin{align}
    \min_F \underset{x}{\E}~  \left[\max_{||\delta||\leq\epsilon} \mathcal{L}(F(x+\delta),y)\right],
    \label{eq:adv_training}
\end{align}
which is the model $F$'s average loss on the data $(x,y)$ that perturbed by adversarial noise $\delta$.
These adversarial training methods~\cite{kurakin2016adversarial, madry2017towards, zhang2019theoretically} have been shown to be one of the most effective approaches to improve the model robustness against adversarial attacks. Note that similar to traditional ERM, adversarially trained models can also achieve very high training performance. For example, under ResNet18~\cite{he2016deep}, PGD adversarial training~\cite{he2016deep} can achieve over 99\% clean accuracy and 85\% adversarial accuracy on the training set of CIFAR~100. Under a larger network with WideResNet28-10 (WRN28), it can achieve 100\% clean accuracy and 99\% adversarial accuracy. It suggests that these DNN models have sufficient capacity to memorize the labels of these atypical samples and their adversarial counterparts. However, different from ERM, adversarially trained models usually suffer from bad generalization performance on the test set. For example, the ResNet18 model can only have 57\% and 22\% test clean accuracy and adversarial accuracy ($\sim59\%$ and $24\%$ on WRN28). Moreover, the study~\cite{rice2020overfitting} suggests that during the adversarial training process, the model's test adversarial accuracy keeps dropping as more training data is fitted (after the first time learning rate decay). Thus, these facts indicate that the memorization in adversarial training is probably not always beneficial to the test performance and requires deep understanding. In the following sections, we will empirically draw a significant connection between these properties with the memorization effect.
\vspace{-0.4cm}
\section{Atypical Samples in Adversarial Training}\label{sec:pre}
\vspace{-0.2cm}
In this section, we attempt to understand adversarial training's behavior by studying its relation with the memorization effect. The discussions are mainly based on PGD-adversarial training~\cite{madry2017towards} on CIFAR~100. Implementation details are shown in Appendix~\ref{app:pre} where we also report the results in more datasets, i.e., CIFAR~10 and Tiny~ImageNet~\cite{le2015tiny}, and we make consistent observations.

\vspace{-0.2cm}
\subsection{Adversarial Robustness of Atypical Samples is Harder to Generalize}\label{sec:pre1}
\begin{figure}[t]
\centering
\hspace*{-1cm}
\subfloat[Clean (left) \& Adv Acc. (right) under ResNet18.]{
\label{fig:harder1}
\begin{minipage}[c]{0.55\textwidth}
\includegraphics[width = 0.5\textwidth]{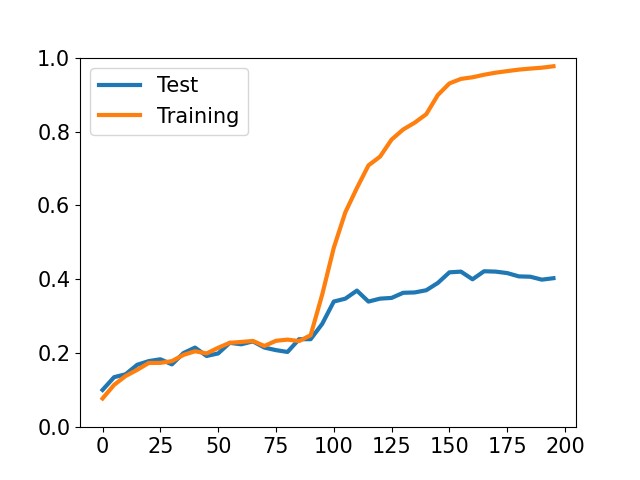}%
\hfill
\includegraphics[width = 0.5\textwidth]{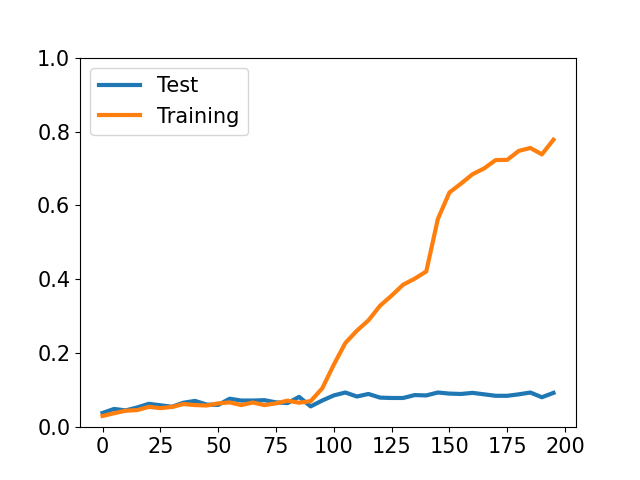}
\end{minipage}
}
\hspace*{-0.4cm}
\subfloat[Clean (left) \& Adv Acc. (right) under WRN28.]{
\label{fig:harder3}
\begin{minipage}[c]{0.55\textwidth}
\includegraphics[width = 0.5\textwidth]{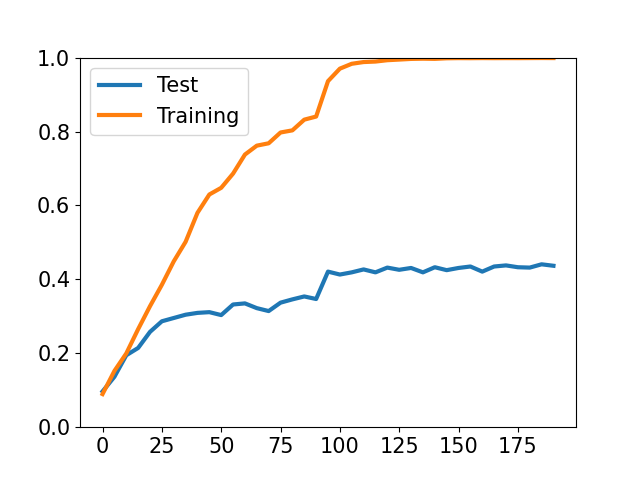}%
\hfill
\includegraphics[width = 0.5\textwidth]{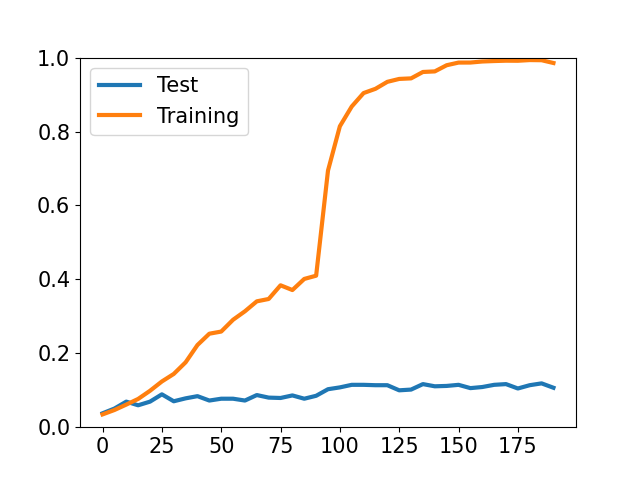}
\end{minipage}
}
\caption{Clean Accuracy and Adversarial Accuracy on \textbf{Atypical} Set of CIFAR100}
\vspace{-0.5cm}
\label{fig:rare_benefit}
\end{figure}
In this subsection, we first check whether fitting atypical samples in adversarial training can effectively help the model correctly and robustly predict the atypical samples in the test set. We apply PGD adversarial training~\cite{madry2017towards} on original CIFAR~100 dataset for 200 epochs and evaluate the model's clean accuracy and adversarial accuracy on training atypical set $\mathcal{D}_\text{atyp}=\{x_i \in \mathcal{D}: \text{mem}(x_i)> 0.15\}$ and its corresponding test atypical set $\mathcal{D}_\text{atyp}' = \{x'_j \in \mathcal{D}': \text{infl}(x_i,x'_j)> 0.15, \text{for } \forall x_i\in \mathcal{D}_\text{atyp}\}$. In Fig.~\ref{fig:rare_benefit}, we report the algorithm's performance (clean \& adv. acc.) on these atypical sets along with the training process. From the results, we observe that both ResNet18 and WRN28 are capable to memorize all clean atypical samples and most adversarial atypical samples, since they both achieve $\approx100\%$ clean accuracy and high adversarial accuracy ($\approx80\%$ and $100\%$, respectively) on the training atypical set. 
As the training goes, the models' clean accuracy on the test atypical set gradually improves and finally approaches 40\%. 
However, their adversarial robustness keep constant around 10\% from the beginning epochs to the last ones, no matter how high the training performance is. 
These results suggest that the memorizing atypical samples in adversarial training may only improve their test clean accuracy, but hardly help their adversarial robustness. Recall that in CIFAR100, atypical set $\mathcal{D}_\text{atyp}$ (with memorization value > 0.15) covers 40\% samples of the whole dataset. Completely failing on the adversarial robustness of atypical samples could be one important reason that contributes to the poor robustness generalization of DNNs~\cite{rice2020overfitting}.

As the previous theoretical study~\cite{schmidt2018adversarially} states, for a model to have good robustness generalization performance, it always demands a training set with much larger amount of samples, than a model to have good clean accuracy generalization. In our case, the sub-population of each particular atypical sample has very low frequency to appear in the training set, and it is always deviated from the main sub-population. Thus, in the sub-population of this atypical sample, it is equivalent to a classification task based on an extremely small dataset, with one or a few training samples given. Therefore, the adversarial robustness of atypical samples can be extremely hard to generalize. 

\vspace{-0.2cm}
\subsection{Memorizing Atypical Samples Hurts Typical Samples' Performance}\label{sec:pre2}
\vspace{-0.2cm}


In this subsection, we further observe that fitting atypical samples will even bring negative effects on ``typical'' samples. Here, we define ``typical'' samples as the subset of training set $\mathcal{D}$ which have low memorization value: $\mathcal{D}_\text{typ} = \{x_i\in\mathcal{D}:\text{mem}(x_i)<0.02\}$. It means that they are not fitted by memorization and are from the main sub-population in their class. To define the test typical set $\mathcal{D}'_\text{typ}$, we exclude all test samples which have high influence values from any atypical training samples, and also exclude the samples that using ERM algorithm $\mathcal{A}$ has low success rate to predict (the samples which cannot be learned from $\mathcal{D}$): $\mathcal{D}'_\text{typ} = \mathcal{D}' - \{x'_j: \text{infl}(x_i,x'_j)> 0.02, \text{for } \forall x_i\in \mathcal{D}_\text{atyp}\} \cup \{x'_j: \text{Pr.}_{F\leftarrow\mathcal{A}(\mathcal{D})}(F(x'_j) = y_j) < 0.8\}$.

\begin{figure}[t]
\centering
\hspace*{-1cm}
\subfloat[Clean (left) \& Adv Acc. (right) under ResNet18.]{
\label{fig:hurt1}
\begin{minipage}[c]{0.55\textwidth}
\includegraphics[width = 0.5\textwidth]{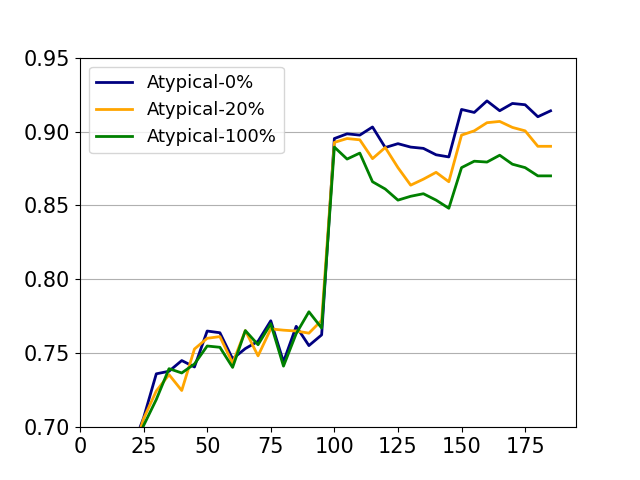}%
\hfill
\includegraphics[width = 0.5\textwidth]{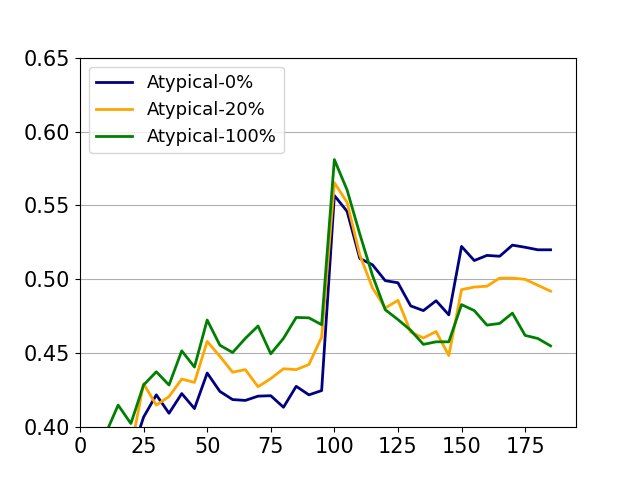}
\end{minipage}
}
\hspace*{-0.4cm}
\subfloat[Clean (left) \& Adv Acc. (right) under WRN28.]{
\label{fig:hurt2}
\begin{minipage}[c]{0.55\textwidth}
\includegraphics[width = 0.5\textwidth]{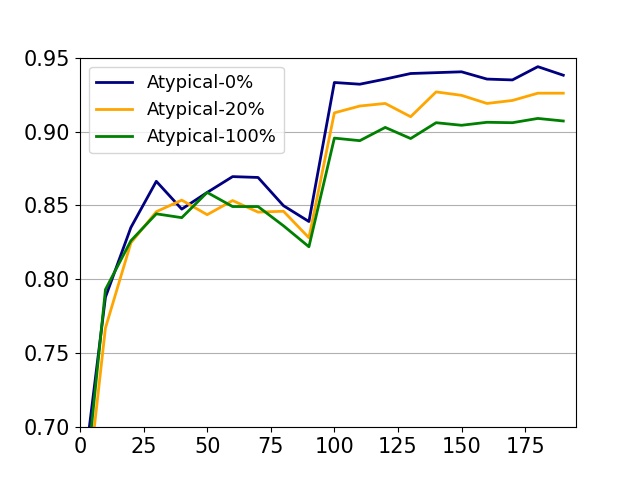}%
\hfill
\includegraphics[width = 0.5\textwidth]{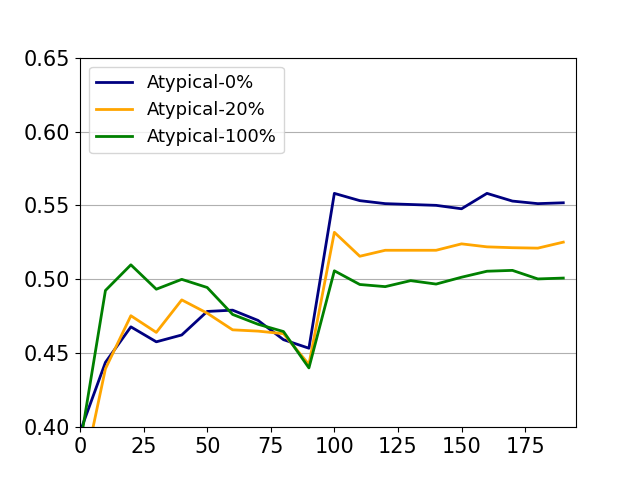}
\end{minipage}
}
\caption{Clean Accuracy and Adversarial Accuracy on \textbf{Typical} Set of CIFAR100}
\vspace{-0.5cm}
\label{fig:hurt}
\end{figure}
To demonstrate the negative effect from fitting atypical samples, we conduct PGD adversarial training~\cite{madry2017towards} for several trails on resampled CIFAR100 datasets: each dataset is constructed with the whole training typical set $\mathcal{D}_\text{typ}$, and a part of the training atypical set $\mathcal{D}_\text{atyp}$ (randomly sample 0\%, 20\% and 100\% in $\mathcal{D}_\text{atyp}$). In Fig.~\ref{fig:hurt}, we report the adversarially trained model's clean and adversarial accuracy on the test typical set $\mathcal{D}_\text{typ}'$ and check the impact of atypical samples on the typical samples. From the results, we find that the amount of atypical samples makes a significant influence on the typical samples. For example, under ResNet18, an adversarially trained model without atypical samples has 92\% clean accuracy and 52\% adversarial accuracy on the test typical samples (on the last epochs). While, the model trained with all atypical samples included only has 85\% and 44\% clean \& adv. accuracy, respectively. 
These results suggest: the more atypical samples exist in training set, the poorer performance the model will have on $\mathcal{D}'_\text{typ}$. In other words, these atypical samples act more like ``poisoning'' samples~\cite{biggio2012poisoning, xu2019adversarial} which can deteriorate the model's performance on typical samples and consequently hurt the overall performance.

\begin{figure}[t]
\subfloat{
\label{fig:poison_pair1}
\begin{minipage}[c]{0.5\textwidth}
\centering
\includegraphics[width = 0.8\textwidth]{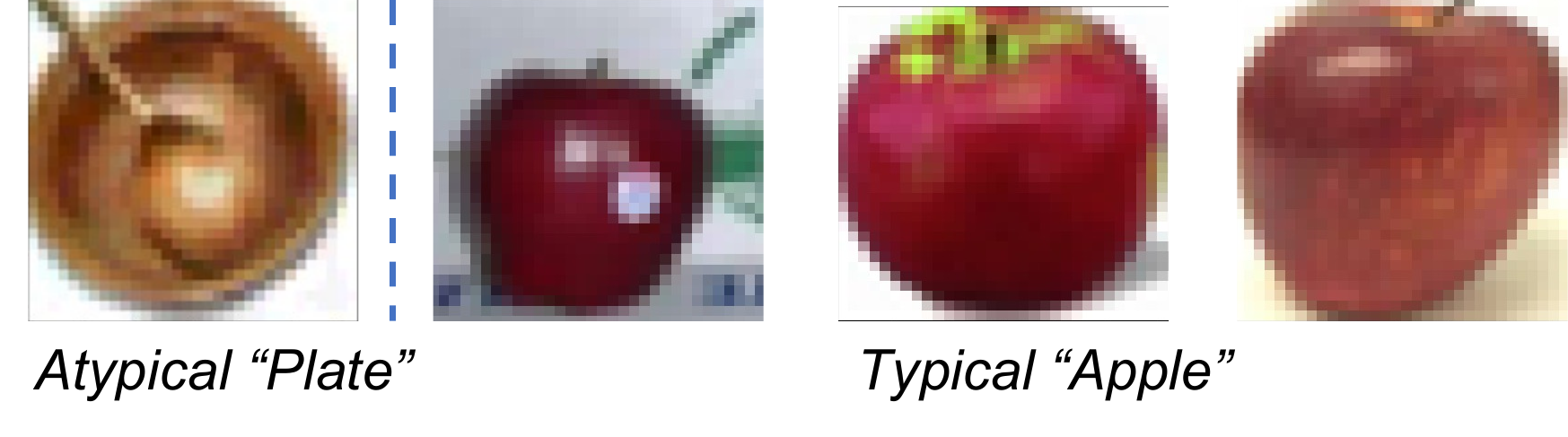}
\end{minipage}
}
\subfloat{\label{fig:poison_pair2}
\begin{minipage}[c]{0.5\textwidth}
\centering
\includegraphics[width = 0.8\textwidth]{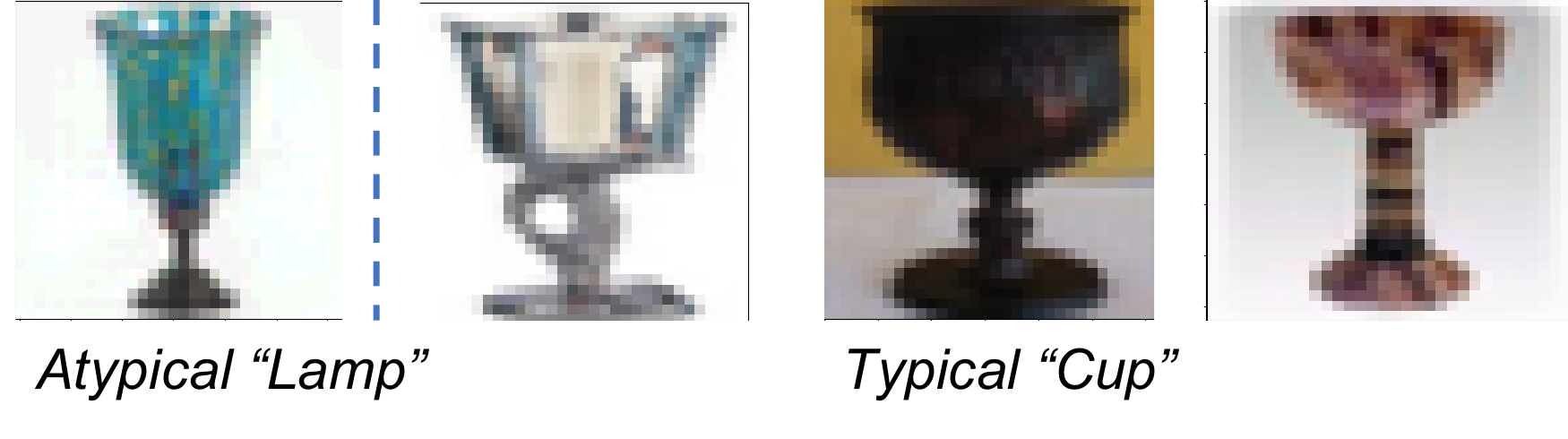}
\end{minipage}
}
\caption{Examples of Poisoning Atypical Samples}
\vspace{-0.4cm}
\label{fig:atypical_samples}
\end{figure}

\noindent\textbf{Poisoning Atypical Samples} A natural question is what kind of atypical samples are likely to ``poison'' model robustness and why? Different from previous literature about poisoning samples in traditional ERM, which assume that poisoning samples are most mis-labeled samples~\cite{li2020gradient}, CIFAR100 is a clean dataset with no or very few wrong labels. However, we hypothesize that the atypical samples which poison the model performance might pertain some features of a ``wrong'' class. Recall that atypical samples are always distinct from the main data distribution in their labeled class, it is likely that they are closer to the distribution of a ``wrong'' class. As shown in  Fig.~\ref{fig:atypical_samples}, an atypical ``plate'' is visually very similar to images in ``apple''. If the model memorizes this atypical ``plate'' and predicts any samples with similar features to be ``plate'', the model cannot distinguish between ``apple'' and ``plate''. As a simple verification to this hypothesis, in Table~\ref{Tab:dist}, we empirically show that atypical samples can cause adversarial training to produce ``less-discriminative'' representations among different classes. Under the same experimental setting and the models above, we measure the average \textit{Cosine Distance (CD)~\footnotemark} of the models' pen-ultimate layer representation output, for each pair of samples (in the training typical set $\mathcal{D}_\text{typ}$) from different classes. Table~\ref{Tab:dist} shows that in adversarial training, fitting more atypical samples will result in a smaller distance for the representations of samples in different classes. It suggests that with atypical samples, DNNs learn more similar and mixed representations for different classes, which can degrade the typical samples' test performance.
\begin{table}[h]
\vspace{-0.5cm}
\small
\centering
\setlength{\tabcolsep}{12pt}
\caption{Class-wise Cosine Distance of Representations of Typical Samples}
\begin{tabular}{c|ccc}
\hline
\# of Atypical Samples & 0\% &20\% &100\%\\
\hline
ResNet18 & 0.66 & 0.62  & 0.59\\
\hline
WRN28 &0.64 & 0.61 & 0.57\\
\hline
\end{tabular}
\vspace{-0.4cm}
\label{Tab:dist}
\end{table}

\footnotetext{Cosine Distance: $\E_{x_1,x_2} [\frac{h(x_1)\cdot h(x_2)}{||h(x_1)||_2\cdot||h(x_2)||_2}]$, where $h(\cdot)$ is the pen-ultimate layer output of DNN model $F(\cdot)$, and $x_1,  x_2\in \mathcal{D}_\text{typ}$ and from different classes.}

It is also worth to mention that this poisoning phenomenon does not appear in the traditional ERM algorithm (results in Appendix~\ref{app:pre}). In ERM, memorizing atypical samples will neither hurt typical sample's performance or degrade the feature space discrimination. A possible explanation is that adversarially trained models use more ``semantically meaningful'' features for prediction~\cite{tsipras2018robustness, ilyas2019adversarial}. During adversarial training, the models will not only memorize the labels of atypical samples, but also their semantic features. As a result, the adversarial trained DNNs will construct more mixed concepts of features, if the ``poisoning'' atypical samples exist.

\vspace{-0.4cm}
\section{Benign Adversarial Training for Memorizing Benign Atypical Samples}
\vspace{-0.2cm}
Based on previous discussions, the effect of atypical samples in adversarial training can be briefly summarized as: 
\textbf{(a)} They can benefit the models' clean accuracy (especially on atypical samples), but hardly improve their robustness. \textbf{(b)} They hurt the performance of typical samples. 
In this section, we ask the question: \textit{Can we eliminate the negative influence from fitting atypical samples during adversarial training?} Admittedly, strategies for adversarial training to stop at early epoch~\cite{rice2020overfitting} is potential to achieve this goal. As shown in Fig.~\ref{fig:hurt1}, the typical samples' performance decreases only after the epoch $100$, which is when the model begins to fit more atypical samples. Stopping at such epoch $100$ can effectively prevent the future performance degradation.
However, it is still unclear how to properly identify such an epoch in general scenarios.
Moreover, the early-stop mechanism tends to ignore all atypical samples and significantly limits the model's ability to handle test atypical samples, especially on complex datasets such as CIFAR100 and ImageNet with large fractions of atypical samples.

In this section, we propose a novel algorithm \textit{Benign Adversarial Training (BAT)}. It is composed of two major components: \textbf{(i) \textit{Reweighting}} to mitigate the impact from those poisoning atypical samples;
\textbf{(ii) \textit{Discrimination Loss}} to further protect the performance on typical samples. Combining these two components, BAT can maintain the performance of typical samples, but still preserve the ability to fit those ``useful''' atypical samples. 
Although BAT can hardly improve the overall robustness compared to adversarial training~\cite{madry2017towards} (when it achieves highest robustness), BAT can present the comparable robustness and higher overall clean accuracy. Compared to~\cite{madry2017towards}, BAT enjoys better clean vs. adversarial accuracy trade-off. More results and discussions can be found in the experiment section. Next, we detail these two components of BAT.

\vspace{-0.2cm}
\subsection{Reweighting \& Poisoning Score} Following PGD adversarial training~\cite{madry2017towards}, BAT starts by fitting manually generated adversarial examples using Projected Gradient Descent (PGD). 
During the training process, in order to identify which samples can poison or degrade the model's performance, we define the ``\textit{poisoning score}'' for each training (adversarial) sample $x_i^\text{adv}$ as:
\begin{align}\label{eq:poisoning_score}
    \textbf{q}(x_i^\text{adv}) = \max_{t\neq y}{F_t(x_i^\text{adv})}
\end{align}
which is the model $F$'s largest prediction score (after softmax) to a wrong class $t$ other than the true class $y$.  A high poisoning score suggests that the current model predicts the sample $x^\text{adv}$ to be a wrong class with high confidence. 
Note that the atypical samples are usually fitted into the model later than typical samples. Under a model with a few atypical samples fitted, if there is an atypical sample $x_i^\text{adv}$ with high $\textbf{q}(x_i^\text{adv})$, $x_i^\text{adv}$ is very likely to be close to the distribution of a wrong class instead of its true class. In Fig.~\ref{fig:atypical_samples}, we present  several atypical (adversarial) samples with  poisoning scores larger than 0.8. These samples present semantic features which are very similar to the wrong class that the model predicts. Therefore, fitting these atypical samples could cause the model to learn mixed concepts of features and degrades the model performance.

\looseness=-1
During the training process, we desire that  the model should mitigate the influence from the atypical samples with large poisoning scores, but still learn other ``useful /benign'' atypical samples. Thus, we design a cost-sensitive reweighting strategy which downweights the cost of atypical samples with large poisoning scores. In particular, we specify the weight value $w_i$ for each training sample $x_i^\text{adv}$ as:
\begin{align}\label{eq:reweight}
w_i = 
\begin{cases}
\exp (-\alpha \cdot \textbf{q}(x_i^\text{adv}))  & \text{~~~if~~~} \text{mem}(x_i)>\sigma \text{~~and~~} \argmax_t F_t(x_i^\text{adv})\neq y\\
1 &\text{~~ Otherwise.}
\end{cases}
\end{align}
where $\alpha\in\R^+$ and $\sigma$ control the size of the reweighted atypical set. Since the function $\exp(-\alpha(\cdot))$ is decreasing and ranges from 1 to 0, the samples with large poisoning scores will be assigned with small weights close to 0.
Thus, in the reweighting algorithm, we train the model to find optimal model $F_\text{rw}$ by assigning the weight vector $w$:
\begin{align}
    F_\text{rw} = \argmin_F  \frac{1}{\sum_i w_i }\sum_i \left[w_i \cdot \mathcal{L}(F(x_i^\text{adv}), y_i)\right]
\end{align}
In the training process, those (adversarial) training samples with large poisoning scores are assigned with small weights and correspondingly their influence will be largely mitigated.

\subsection{Discrimination Loss.}
Even though the reweighting method discussed above can help mitigate the impact from those ``poisoning'' atypical samples, we observe that the model's performance (especially on the typical samples) is still inevitably decreasing with training goes. For example, as shown in Fig.~\ref{fig:hurt}, even we exclude all atypical samples in CIFAR100 dataset, the typical samples' clean and adversarial accuracy still slightly drop. Fortunately, motivated from the discussions in Section~\ref{sec:pre2}, the performance of the typical samples is strongly related to the distance of the representation vectors of different classes. It means the smaller representation distance among different classes, the poorer discrimination the model has to distinguish different classes. 
Thus, to further preserve the model's performance on typical samples, we introduce \textit{Discrimination Loss} to regularize the feature space discrimination for typical samples among different classes:
\begin{align}
\label{eq:dl}
\begin{split}
\mathcal{L}_{DL} (F) = &
\underset{{\{(x_k, y_k)\}^K_{k=1}}}{
\underset{(x_i,y_i), (x_j, y_j)} {\E}  }
\left[- \log ~~\frac{e^{h(x_i^\text{adv}) \cdot h(x_j^\text{adv}) / \tau}}{ \sum_{k=1}^K  e^{h(x_i^\text{adv}) \cdot h(x_k^\text{adv}) / \tau}} \right].\\
\text{where~~~}  & y_i= y_j; \text{~~~} y_k\neq y_i, \text{~}\forall k=1,2,...K \\
&\text{mem}(x_i), \text{mem}(x_j), \text{mem}(x_k) < \sigma\\
\end{split}
\end{align}
Specifically, $h(\cdot)$ is the model $F$'s pen-ultimate layer's output representation; $\tau\in\R^+$ is the temperature value and $K\in \mathbb{Z}^+$ is a fixed number of the samples that randomly chosen with different labels with $y_i$\footnotemark.
Intuitively, constraining the \textit{Discrimination Loss} imposes the model to output similar representations for the typical sample pairs $(x_i^\text{adv}, x_j^\text{adv})$ in the same class, and distinct representations for $(x_i^\text{adv}, x_k^\text{adv})$ from different classes. Similar approaches have been widely used in
representation learning, such as Triplet Loss~\cite{schroff2015facenet} and contrastive learning methods~\cite{chen2020simple}, which stress the good property of DNNs' representation space.
With the Discrimination Loss incorporated, the final objective of BAT is formulated as:
\begin{align}
    F_\text{bat} =\argmin_F  \left(\frac{1}{\sum_i w_i }\sum_i \left[w_i \cdot \mathcal{L}(F(x_i^\text{adv}), y_i)\right] + \beta\cdot\mathcal{L}_{DL}(F) \right)
\end{align}
where $\beta>0$ that controls the intensity of Discrimination Loss.
The algorithm scheme of Benign Adversarial Training is presented in Appendix~\ref{app:algorithm}.

\footnotetext{In practice, under training algorithms using mini-batches, the set $\{x_k\}^K_{k=1}$ can be replaced by all (typical) samples in the mini-batch, with different labels from $y_i$.}

\vspace{-0.2cm}
\section{Experiment}\label{sec:exp}
\vspace{-0.2cm}
In this section, we present the experimental results to validate the effectiveness of the proposed BAT algorithm on benchmark datasets and compare it with state-of-the-art baseline methods. We also verify that both two components are helpful and necessary for BAT. The implementation of the BAT can be found via \url{https://anonymous.4open.science/r/benign-adv-77C5}.

\vspace{-0.2cm}
\subsection{Experimental Setup}
\vspace{-0.1cm}
In this work, in order to demonstrate the 
merit of BAT, we conduct the experiments mainly on benchmark datasets CIFAR100~\cite{krizhevsky2009learning} and Tiny~ImageNet~\cite{le2015tiny}, which are relatively complex datasets (i.e., containing larger fractions of atypical samples). For both datasets, we study the algorithms under the model architectures ResNet and WideResNet (WRN)~\cite{he2016deep}. In this section, we only present the results of ResNet18 for CIFAR100 and ResNet32 for Tiny~ImageNet and leave the results on WRN in Appendix~\ref{app:exp}. As a fair comparison with BAT, we implement the baseline algorithms including PGD adversarial training~\cite{madry2017towards} as well as its most popular variant TRADES~\cite{zhang2019theoretically}. In addition, we include several recent algorithms: MART~\cite{wang2019improving} and GAIRAT~\cite{zhang2020geometry}, which also incorporate reweighting strategies into adversarial training. For BAT and all baseline methods, we run the algorithms using SGD~\cite{bottou2010large} for 160 epochs with the learning rate that starts from 0.1 and decays by 0.1 after the epoch 80 and 120. More implementation details can be found in Appendix~\ref{app:exp}.

\textbf{Performance on CIFAR100.} For a comprehensive comparison between different methods on CIFAR100, in the results from Table~\ref{Tab:results_cifar100}, we report the models' clean accuracy and adversarial accuracy against $l_\infty$-$8/255$ PGD attack~\cite{madry2017towards}, as well as their performance on the typical sample set and atypical sample set (as described in Section~\ref{sec:pre}). A more comprehensive robustness evaluation on different attacking methods (including CW~\cite{carlini2017towards} and Auto-Attack~\cite{croce2020reliable}) are presented in Appendix~\ref{app:exp}. For BAT, we report its performance when choosing its optimal hyperparameter: $\alpha = 1  ~\text{\&} ~2$ and $\beta = 0.2$. In the Section~\ref{sec:ablation}, we will discuss the impact of the selection of $\alpha$ and $\beta$ on BAT. For baseline methods, the settings and checkpoint selections follow the original papers' suggestions.


\vspace{-0.3cm}
\begin{table}[h]
\small
\centering
\caption{Performance of BAT vs. Baselines on CIFAR100 Under ResNet18}
\begin{tabular}{c|cc|cc|cc}
\hline
Method & All Acc. & All Adv. & Typical Acc. & Typical Adv. & Atyp. Acc. & Atyp. Adv. \\
\hline
\hline
PGD Train (Best Adv.) & 56.9 & 27.4 & 90.6 & 59.0 &29.5 & 7.7\\
PGD Train (Best Clean) & 57.8 & 21.9 & 88.3 & 51.0 & \textbf{40.1} & 8.3 \\
TRADES ($1/\lambda = 5$) & 56.6 & 26.9 & 88.9 & 57.1 & 37.3 & \textbf{10.9} \\
MART~\cite{wang2019improving} & 51.8 & \textbf{30.4} & 85.3 & \textbf{62.2} & 25.3 & 10.1\\
GAIRAT~\cite{zhang2020geometry} & 58.2 & 27.8 & 90.6 & 60.7 & 31.6 & 8.2\\
\hline
BAT ($\alpha = 1, \beta = 0.2$) & \textbf{59.5} &27.3 & \textbf{92.3} & 58.8 & 36.3 & 8.7 \\
BAT ($\alpha = 2, \beta = 0.2$) &59.3 & 27.4 & \textbf{92.3} & 60.3 & 33.1 & 7.4 \\
\hline
\hline
\end{tabular}
\label{Tab:results_cifar100}
\end{table}
\vspace{-0.2cm}
From the results in Table~\ref{Tab:results_cifar100}, we can find that BATs enjoy good clean \& adversarial accuracy trade-off among all methods. It is because BATs can obtain the highest overall clean accuracy ($\sim59.5\%$), as well as comparably good adversarial accuracy ($\sim27.3$) with baseline methods. The only exception is MART~\cite{wang2019improving} which has the higher adversarial accuracy around $\sim 30\%$. However, MART~\cite{wang2019improving} has much lower clean accuracy than BATs. 

To gain a deeper understanding of the working mechanism of BAT, we focus on the performance on BAT vs. PGD adversarial training~\cite{madry2017towards}. Compared with PGD adversarial training with highest adversarial accuracy (PGD Train (Best Adv.)), the BAT methods are $2\sim3\%$ higher in terms of overall clean accuracy, and similar adversarial accuracy. The improvement of clean accuracy is mainly due to BATs' capacity to fit much more atypical samples. For example, the clean accuracy of BAT ($\alpha=1, \beta=0.2$) is about $6\%$ higher than PGD Train~(Best Adv.). On the other hand, compared to PGD adversarial training with the highest clean accuracy (PGD Train (Best Clean)), BATs have much better overall adversarial robustness and slightly better clean accuracy. This is mainly due to BAT's advantage on typical samples, because BATs can successfully prevent the performance drop of typical samples during the training process. Since BATs can achieve good performance on both typical samples and atypical samples, BATs outperform PGD adversarial training in CIFAR100.

\begin{wrapfigure}{r}{0.6\textwidth}
\vspace{-0.4cm}
\subfloat{
\begin{minipage}[c]{0.3\textwidth}
\centering
\includegraphics[width = 1.1\textwidth]{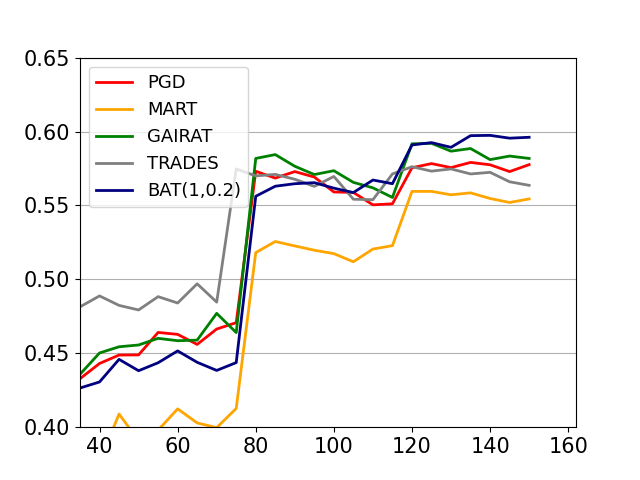}
\end{minipage}
}
\subfloat{
\begin{minipage}[c]{0.3\textwidth}
\centering
\includegraphics[width = 1.1\textwidth]{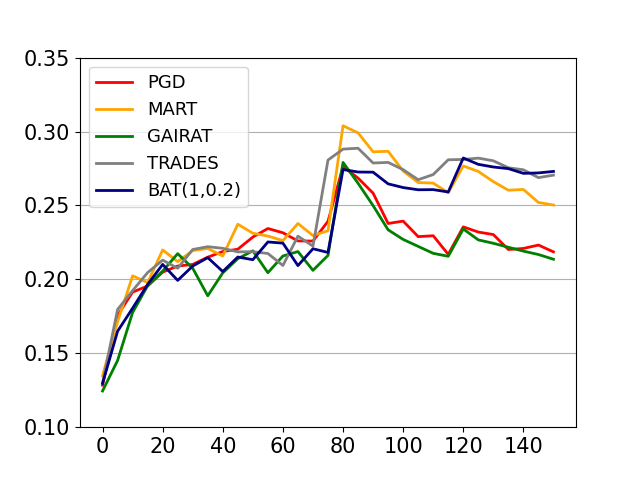}
\end{minipage}
}
\vspace{-0.4cm}
\caption{\small Clean Acc. (left) \& Adv. Acc (right)}
\label{fig:exp1}
\end{wrapfigure}
In Fig.~\ref{fig:exp1}, we also show the change of each models' overall clean accuracy (Fig~\ref{fig:exp1} (left)) \& adversarial accuracy (Fig~\ref{fig:exp1} (right)) along with the training progress. 
Interestingly, similar to PGD adversarial training, all baseline methods cannot achieve the optimal clean and adversarial accuracy at the same moment.
They always achieve the best adversarial accuracy around Epoch 80 (right after the first time weight decay), and the best clean accuracy on the last epochs. However, for BATs, since they can effectively prevent the robustness dropping in the late epochs, BATs are able to train until last epochs and enjoy good clean and adversarial accuracy simultaneously.

\textbf{Performance on Tiny~ImageNet.} Tiny~ImageNet~\cite{le2015tiny} contains 200 classes of the images in the original ImageNet~\cite{krizhevsky2012imagenet} dataset, with 500 training images for each class, and image size $64\times 64$. In our experiments, we only choose the first 50 classes in Tiny ImageNet for training and prediction. Since the image size is $64\times 64$, for both training and robustness evaluation, we consider the adversarial attacks are bounded by $l_\infty$-norm-4/255. In Table~\ref{Tab:results_imagenet}, we report the performance of BAT and baseline methods. Similar to the conclusions we can make from CIFAR100, BATs can achieve the highest overall clean accuracy and comparably good adversarial accuracy with baseline methods. 
\begin{table}[h]
\small
\vspace{-0.2cm}
\centering
\caption{Performance of BAT vs. Baselines on Tiny~ImageNet Under ResNet32}
\begin{tabular}{c|cc|cc|cc}
\hline
Method & All Acc. & All Adv. & Typical Acc. & Typical Adv. & Atyp. Acc. & Atyp. Adv. \\
\hline
\hline
Adv. Train (Best Adv.) & 56.3 & 32.3 & 97.5 & 85.3 &41.5 & 9.6\\
Adv. Train (Best Clean) & 58.2 & 30.5 & 98.0 & 80.4 & 44.7 & 9.1 \\
TRADES ($1/\lambda = 5$) & 55.4 & 28.8 & 97.3 & 77.4 & 38.8 & 9.6 \\
MART~\cite{wang2019improving} & 56.2 & \textbf{34.5} & 97.7 & 85.1 & 41.4 & \textbf{13.6} \\
GAIRAT~\cite{zhang2020geometry} & 58.4 & 30.4 & 98.0 & 81.7 & 45.7 & 7.8\\
\hline
BAT ($\alpha = 1, \beta = 0.2$) & \textbf{59.4} &32.0 & 98.4 & 83.9 & \textbf{48.4} & 10.2 \\
BAT ($\alpha = 2, \beta = 0.2$) &\textbf{59.4} &32.9 & \textbf{99.1} & \textbf{86.9} & 45.7 & 10.9 \\
\hline
\hline
\end{tabular}
\label{Tab:results_imagenet}
\end{table}

\vspace{-0.7cm}
\subsection{Ablation Study}\label{sec:ablation}
\vspace{-0.2cm}
\begin{wrapfigure}{r}{0.6\textwidth}
\vspace{-0.9cm}
\subfloat{
\begin{minipage}[c]{0.32\textwidth}
\centering
\includegraphics[width = 1.1\textwidth]{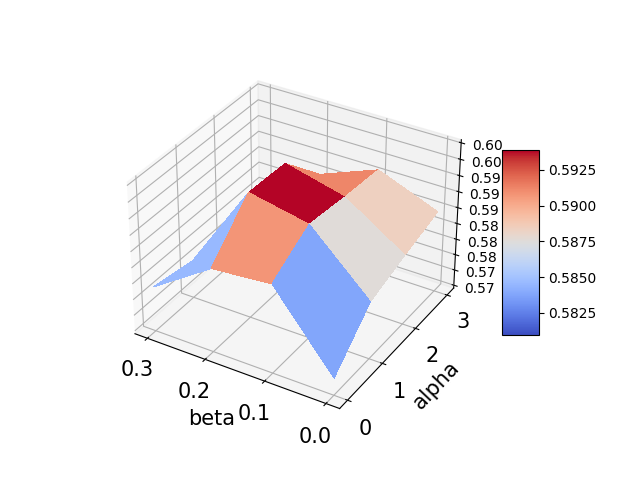}
\end{minipage}
}
\subfloat{
\begin{minipage}[c]{0.32\textwidth}
\centering
\includegraphics[width = 1.1\textwidth]{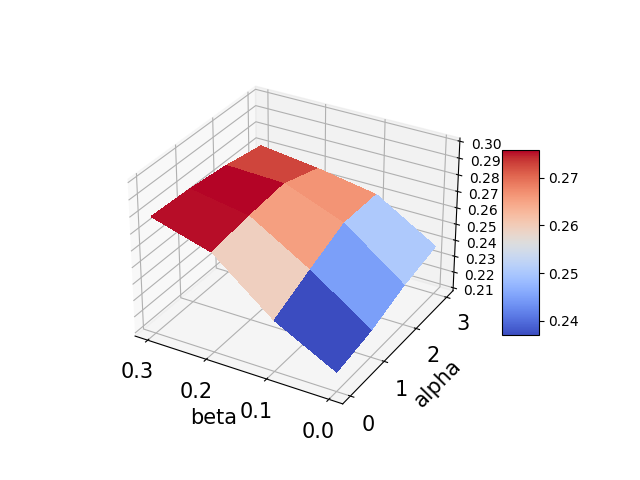}
\end{minipage}
}
\caption{\small Clean Acc.(left) \& Adv. Acc.(right)}
\label{fig:exp2}
\vspace{-0.2cm}
\end{wrapfigure}
In this subsection, we study the potential impact of the hyperparameters chosen in BAT, which is $\alpha$ that controls the \textit{Reweighting} process and $\beta$ that controls the \textit{Discrimination Loss}. 
In Fig.~\ref{fig:exp2}, we conduct the experiments on CIFAR100 with BAT when $\alpha$ is chosen in [0,1,2,3] and $\beta$ is in [0, 0.1, 0.2, 0.3].
In Fig.~\ref{fig:exp2}, we show each model's overall clean accuracy~(left) \& adversarial accuracy~(right) along the Z-axis, and X-axis / Y-axis indicate the models' corresponding variables of $(\alpha,\beta)$. Note that when $\alpha=\beta=0$, the BAT method regresses to original PGD adversarial training. From the result, we can see that a positive pair of $(\alpha,\beta)$ can benefit both model clean and adversarial accuracy. Therefore, both two components \textit{Reweighting} and \textit{Discrimination Loss} of BAT are helpful and necessary. However, when $\alpha$ or $\beta$ is relatively too large, it will hurt the BAT's clean accuracy. As a result, in CIFAR~100, when $\alpha=1$ or 2 and $\beta= 0.2$, BAT can achieve the optimal performance.


\vspace{-0.4cm}
\section{Related Works}
\vspace{-0.3cm}
\textbf{Memorization and atypical samples.} The memorization effect of overparameterized DNNs have been extensively studied both empirically~\cite{zhang2016understanding, nakkiran2019deep} and theoretically~\cite{bartlett2002rademacher}. From traditional views, the memorization can be harmful to the model generalization, because it makes DNN models easily fit those outliers and noisy labels.  However, recent studies point out the concept of ``benign overfitting''~\cite{bartlett2020benign, feldman2020does, feldman2020neural}, which suggests the memorization effect necessary for DNNs to have extraordinary performance on modern machine learning tasks. 
Especially, the recent work~\cite{feldman2020neural} empirically figures out those atypical/rare samples in benchmark datasets and show the contribution from memorizing atypical samples to the DNN's performance. Besides the work~\cite{feldman2020neural}, there are also other strategies~\cite{carlini2019distribution} to find atypical samples in training dataset.  Notably, our work is not the first effort to study the influence of memorization on DNN's adversarial robustness. A previous study~\cite{sanyal2020benign} illustrates that memorizing the mis-labeled samples might be a reason to cause the DNNs' adversarial vulnerability. In our paper, we focus on atypical samples, which appear much more frequently in common datasets, and we study their impacts especially on adversarial training algorithms~\cite{madry2017towards, zhang2019theoretically,chatterji2020finite, muthukumar2020harmless}.\\
\textbf{Adversarial robustness. } 
Adversarial training methods~\cite{madry2017towards, zhang2019theoretically, wang2019improving, zhang2016understanding, rice2020overfitting} are considered as one of the most reliable and effective methods to protect DNN models against adversarial attacks~\cite{goodfellow2014explaining, xu2019adversarial}. However, there are several intrinsic properties of adversarial training which requires deeper understandings. 
For example, they always suffer from poor robustness generalization~\cite{ schmidt2018adversarially, rice2020overfitting}, and they always present strong trade-off relation between clean accuracy vs. robustness~\cite{tsipras2018robustness, zhang2019theoretically}. Our work aims to study these properties from the data perspective and demonstrate the significant connection of the memorization effect with these properties.

\vspace{-0.4cm}
\section{Conclusion}
\vspace{-0.2cm}
In this paper, we draw significant connections of the memorization effect of deep neural networks with the behaviors of adversarial training algorithms. Based on the findings, we devise a novel algorithm BAT to enhance the performance of adversarial training. The findings of the paper can motivate the futures studies in building robust DNNs with more attention on the data perspective.
\bibliographystyle{unsrt}
\bibliography{sample}

\appendix
\newpage
\begin{center}
\large{\bf{Appendix}}
\end{center}

In the supplementary materials, we provide more details about atypical samples and the proposed algorithm, as well as the full experimental results of the preliminary study and the proposed method. 
\begin{itemize}
    \item Additional Introduction of Atypical Samples\hfill Appendix~\ref{app:atypical}
    \item Additional Results of Preliminary Study (in Section~\ref{sec:pre1})\hfill Appendix~\ref{app:pre1}
    \subitem Traditional ERM \& Adversarial Training
    \item Additional Results of Preliminary Study (in Section~\ref{sec:pre2})\hfill Appendix~\ref{app:pre2}
    \subitem Traditional ERM \& Adversarial Training
    \item Full Training Scheme of BAT\hfill Appendix~\ref{app:algorithm}
    \item Additional Results of BAT\hfill Appendix~\ref{app:exp}
    \item Boarder Impacts of this Paper \hfill Appendix~\ref{app:board}
\end{itemize}

\section{Additional Introduction of Atypical Samples}\label{app:atypical}

\begin{figure}[h]
    \centering
    \includegraphics[width = 0.9\linewidth]{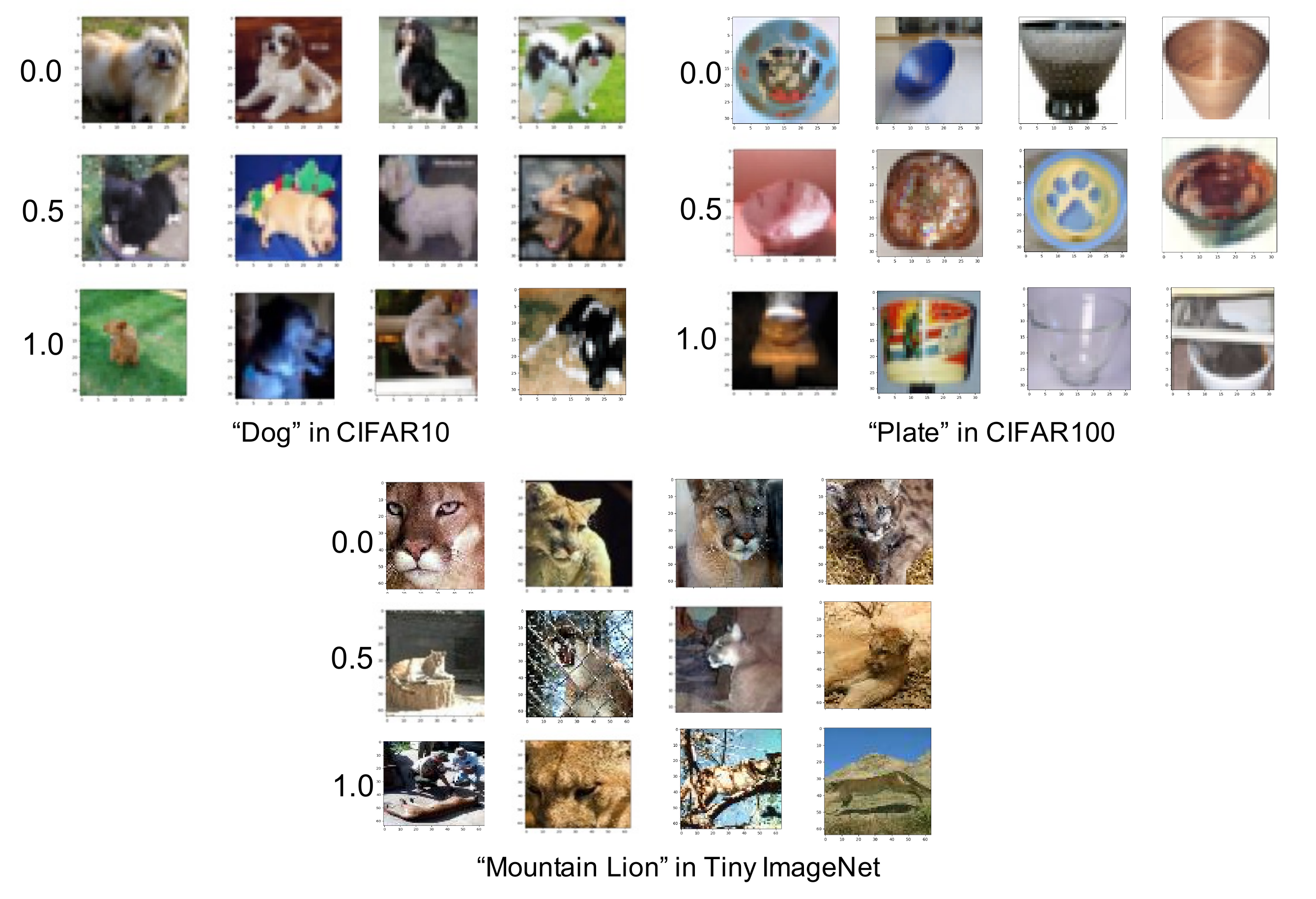}
    \caption{Examples of Images with Different Memorization Values}
    \label{fig:show_mem}
\end{figure}

In this section, we provide additional introductions about the atypical samples in common datasets.
In Fig.~\ref{fig:show_mem}, we provide several examples of images from CIFAR10, CIFAR100~\cite{krizhevsky2009learning} and Tiny ImageNet~\cite{le2015tiny} respectively, with different memorization value (as defined in Section~\ref{sec:def_atypical}) around $0.0, 0.5, 1.0$. These examples suggest that if the memorization value of an image is large, this image is very likely to be ``atypical'', as it presents very distinct semantic features with the images in the main distribution (with memorization value 0.0). The detailed introduction about how to estimate the memorization value in practice can be found in the  work~\cite{feldman2020neural}.

\begin{figure}[h]
\centering
\subfloat[CIFAR100.]{
\begin{minipage}[c]{0.3\textwidth}
\includegraphics[width = 1.0\textwidth]{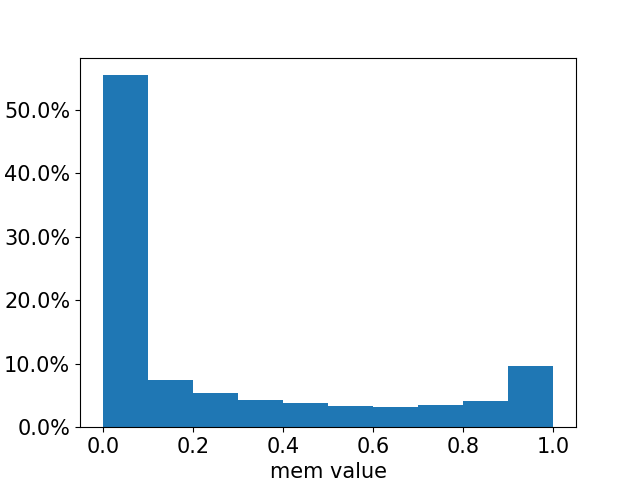}
\end{minipage}
}
\hspace*{0.0cm}
\subfloat[CIFAR10.]{
\begin{minipage}[c]{0.3\textwidth}
\includegraphics[width = 1.0\textwidth]{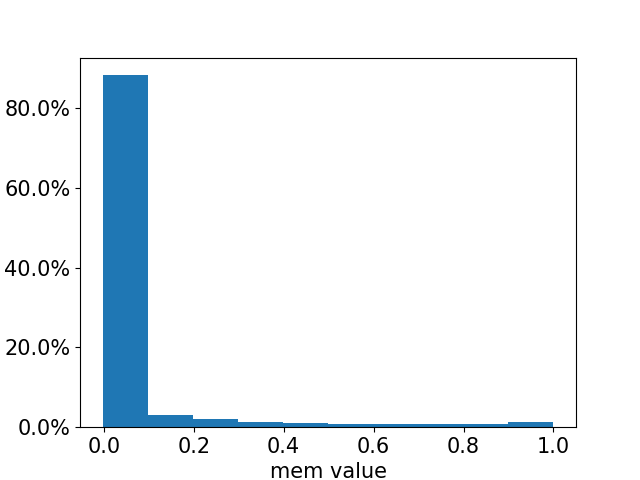}%
\end{minipage}
}
\hspace*{0.0cm}
\subfloat[Tiny~ImageNet.]{
\begin{minipage}[c]{0.3\textwidth}
\includegraphics[width = 1.0\textwidth]{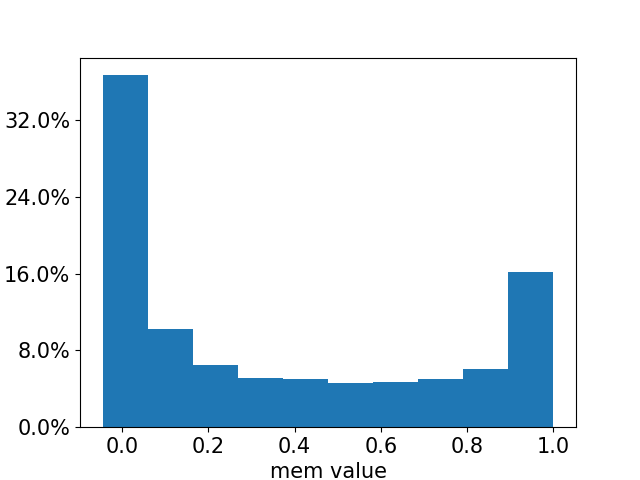}%
\end{minipage}
}
\caption{Frequencies of Training samples with Different \textbf{Memorization Values} in Various Datasets}
\label{fig:hist}
\end{figure}

\newpage
In Fig.~\ref{fig:hist}, we provide histograms to show the distribution of the estimated memorization values of all training samples from CIFAR10, CIFAR100 and Tiny~ImageNet. From Fig.~\ref{fig:hist}, we can observe that atypical samples (with high memorization value > 0.15) consist of a significant fraction (over 40\% \& 50\% respectively) in CIFAR100 and Tiny~ImageNet. In CIFAR10, they also consist of a non-ignorable fraction which is over 10\%.

\begin{figure}[h]
    \centering
    \includegraphics[width = 0.75\linewidth]{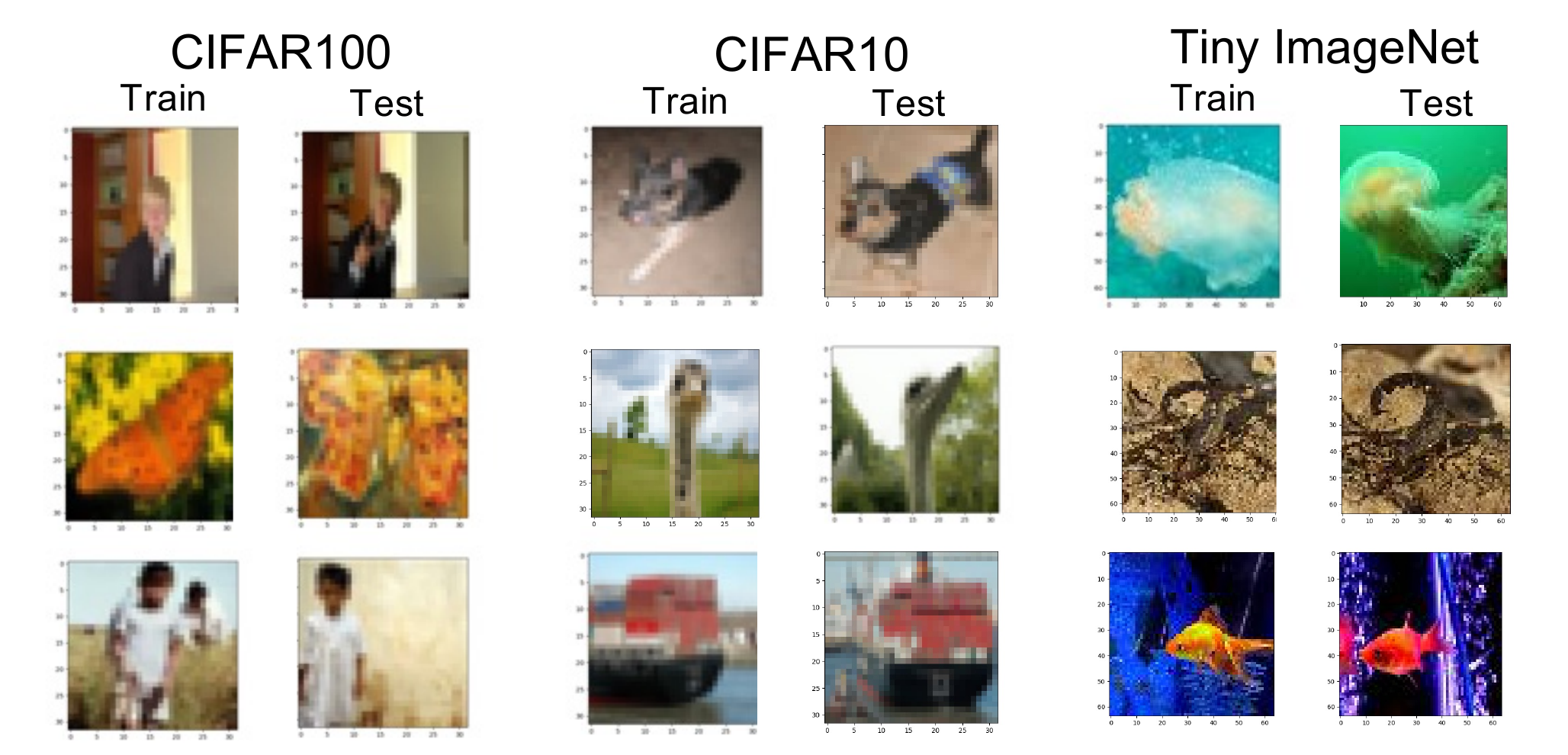}
    \caption{High Influence Pairs with Influence Value > 0.15}
    \label{fig:show_pair}
\end{figure}

In Fig.~\ref{fig:show_pair}, we provide several pairs of images with high influence value (as defined in Section~\ref{sec:def_atypical}) which is over 0.15. In each pair, the training sample also has a high memorization value over 0.15. These examples suggest that there exist atypical samples in both training \& test sets of CIFAR10, CIFAR100 and Tiny~ImageNet. A pair of atypical samples (in the training set and test set) with a high influence value are visually very similar. Moreover, since they have high influence values, removing the atypical samples in the training set is very likely to cause the model to fail on the test atypical samples. Therefore, without memorizing the atypical sample in the training set, the model can hardly predict the atypical samples in the test set.

\section{Additional Results for Preliminary Study}\label{app:pre}

In this section, we provide the full results of the preliminary study in Section~\ref{sec:pre} 
on CIFAR10, CIFAR100 and Tiny~ImageNet, to illustrate the distinct behaviors of the memorization effect between traditional ERMs and adversarial training. In both ERM and adversarial training, we train the models under ResNet18 and WideResNet28-10 (WRN28) architectures. In the experiments, we train the models for 200 epochs with learning rate 0.1, momentum 0.9, weight decay 5e-4, and decay the learning rate by 0.1 at the epoch 150 and 200. For adversarial training, we conduct experiments using PGD adversarial training~\cite{madry2017towards} by default to defense against $l_\infty$-8/255 adversarial attack, with the exception on Tiny~ImageNet, which is against $l_\infty$-4/255 attack. For robustness evaluation, we conduct a 20-step PGD attack.

\subsection{Additional Results for Preliminary Study - Section~\ref{sec:pre1}}\label{app:pre1}

In this subsection, we provide more experimental results to validate the statement 
in Section~\ref{sec:pre1}, where we state that fitting atypical samples in adversarial training can only improve the clean accuracy of test atypical samples, but hardly help their adversarial robustness. We provide full empirical results to show that: \textbf{(i)} In traditional ERM, fitting atypical samples improves the clean accuracy of test atypical samples. \textbf{(ii)} In adversarial training, fitting (adversarial) atypical samples improves the clean accuracy of test atypical samples but can hardly improve the adversarial robustness of them. The experimental setting follows Section~\ref{sec:pre1}, where we apply traditional ERM and adversarial training on original CIFAR10, CIFAR100, Tiny~ImageNet datasets. We evaluate the model's clean accuracy and adversarial accuracy on training atypical set $\mathcal{D}_\text{atyp}=\{x_i \in \mathcal{D}: \text{mem}(x_i)> 0.15\}$ and its corresponding test atypical set $\mathcal{D}_\text{atyp}' = \{x'_j \in \mathcal{D}': \text{infl}(x_i,x'_j)> 0.15, \text{for } \forall x_i\in \mathcal{D}_\text{atyp}\}$.

\textbf{(i) Additional Results for Preliminary Study - Section~\ref{sec:pre1} In Traditional ERM}

\begin{figure}[h]
\centering
\hspace*{-1cm}
\subfloat[ResNet18.]{
\begin{minipage}[c]{0.3\textwidth}
\includegraphics[width = 1.0\textwidth]{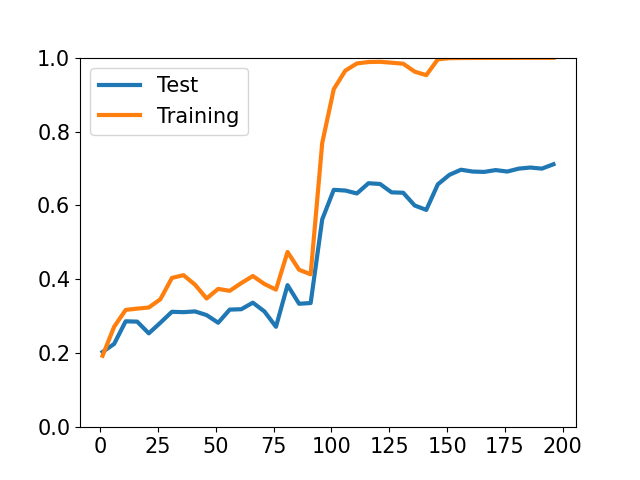}
\end{minipage}
}
\hspace*{0.4cm}
\subfloat[ WRN28.]{
\begin{minipage}[c]{0.3\textwidth}
\includegraphics[width = 1.0\textwidth]{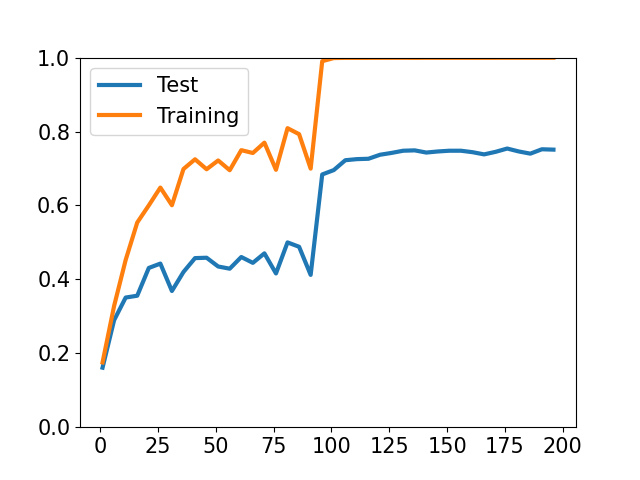}%
\end{minipage}
}
\caption{Clean Accuracy on \textbf{Atypical} Set of CIFAR100}
\label{fig:app_1_11}
\vspace{-0.5cm}
\end{figure}

\begin{figure}[h]
\centering
\hspace*{-1cm}
\subfloat[ ResNet18.]{
\begin{minipage}[c]{0.3\textwidth}
\includegraphics[width = 1.0\textwidth]{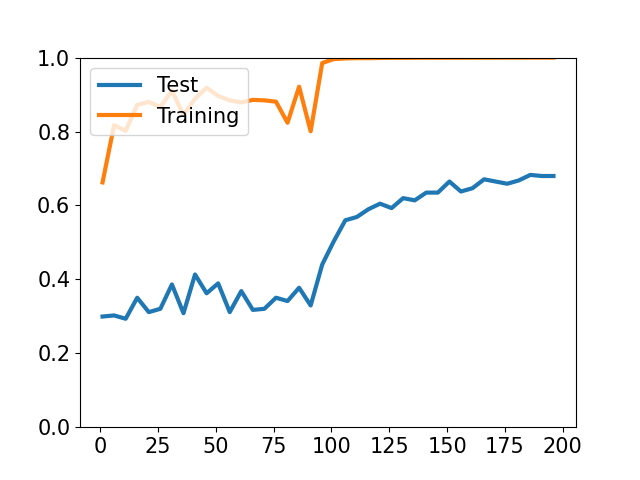}
\end{minipage}
}
\hspace*{0.4cm}
\subfloat[WRN28.]{
\begin{minipage}[c]{0.3\textwidth}
\includegraphics[width = 1.0\textwidth]{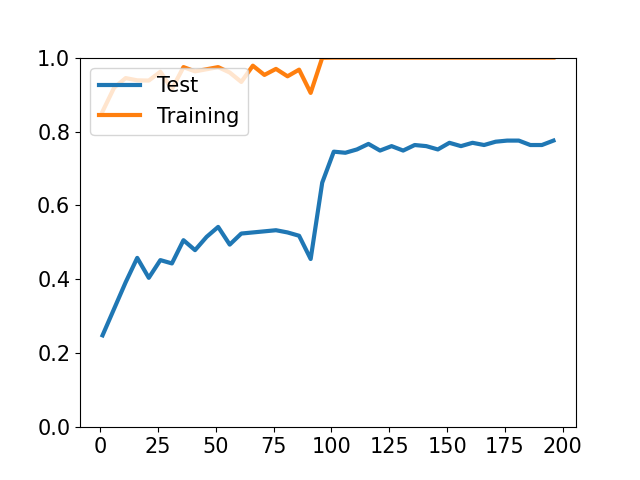}%
\end{minipage}
}
\caption{Clean Accuracy on \textbf{Atypical} Set of CIFAR10}
\label{fig:app_1_12}
\vspace{-0.5cm}
\end{figure}

\begin{figure}[h!]
\centering
\hspace*{-1cm}
\subfloat[ResNet32.]{
\begin{minipage}[c]{0.3\textwidth}
\includegraphics[width = 1.0\textwidth]{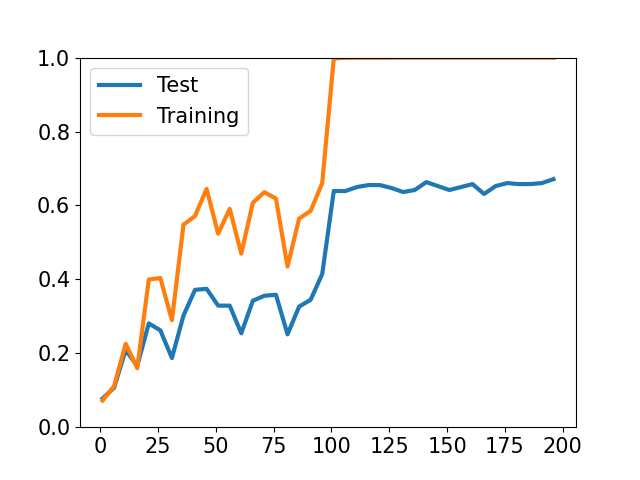}
\end{minipage}
}
\hspace*{0.4cm}
\subfloat[WRN28.]{
\begin{minipage}[c]{0.3\textwidth}
\includegraphics[width = 1.0\textwidth]{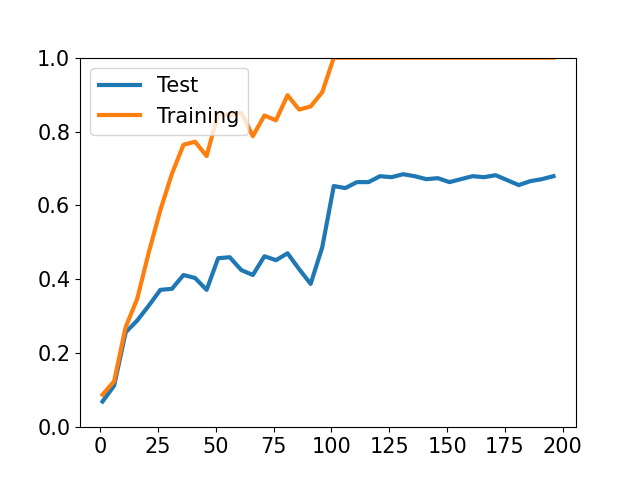}%
\end{minipage}
}
\caption{Clean Accuracy on \textbf{Atypical} Set of Tiny~ImageNet}
\label{fig:app_1_13}
\end{figure}

Fig.~\ref{fig:app_1_11}, Fig.~\ref{fig:app_1_12} and Fig.~\ref{fig:app_1_13} report the performance (clean accuracy) of traditional ERM, which is evaluated on atypical sets under ResNet18 (left) and WRN28 (right) on CIFAR100, CIFAR10 and Tiny~ImageNet. From the figures, we can obverse that fitting atypical samples during training can effectively help the models to achieve good clean accuracy on test atypical samples in all datasets. Note that here we only report clean accuracy as they are not robust against adversarial attacks.

\newpage
\textbf{(ii) Additional Results for Preliminary Study - Section~\ref{sec:pre1} In Adversarial Training}

\begin{figure}[h]
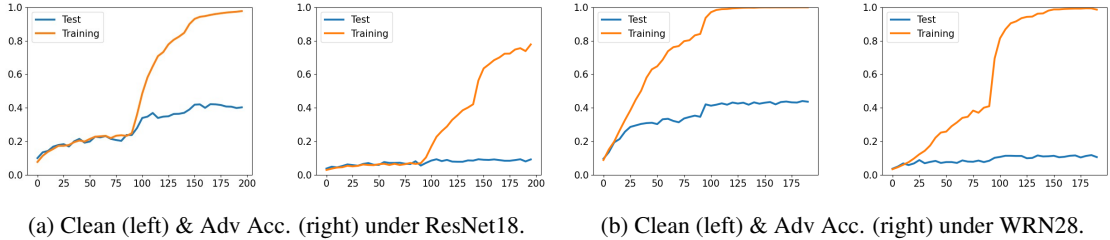

\centering
\hspace*{-1cm}
\subfloat[Clean (left) \& Adv Acc. (right) under ResNet18.]{
\begin{minipage}[h]{0.55\textwidth}
\includegraphics[width = 0.5\textwidth]{figures/clean_rare_cifar.jpg}%
\hfill
\includegraphics[width = 0.5\textwidth]{figures/adv_rare_cifar100.png}
\end{minipage}
}
\hspace*{-0.4cm}
\subfloat[Clean (left) \& Adv Acc. (right) under WRN28.]{
\begin{minipage}[c]{0.55\textwidth}
\includegraphics[width = 0.5\textwidth]{figures/wrn_clean_rare_cifar100.png}%
\hfill
\includegraphics[width = 0.5\textwidth]{figures/wrn_adv_rare_cifar100.png}
\end{minipage}
}
\caption{Clean Accuracy and Adversarial Accuracy on \textbf{Atypical} Set of CIFAR100}
\label{fig:app_1_21}
\vspace{-0.5cm}
\end{figure}

\begin{figure}[h]
\centering
\hspace*{-1cm}
\subfloat[Clean (left) \& Adv Acc. (right) under ResNet18.]{
\begin{minipage}[h]{0.55\textwidth}
\includegraphics[width = 0.5\textwidth]{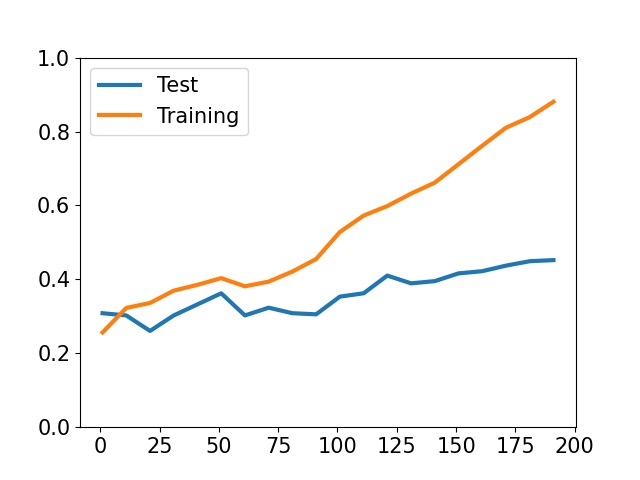}%
\hfill
\includegraphics[width = 0.5\textwidth]{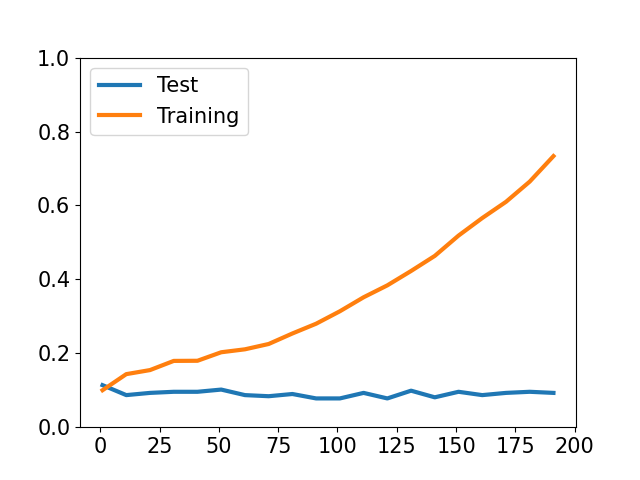}
\end{minipage}
}
\hspace*{-0.4cm}
\subfloat[Clean (left) \& Adv Acc. (right) under WRN28.]{
\begin{minipage}[c]{0.55\textwidth}
\includegraphics[width = 0.5\textwidth]{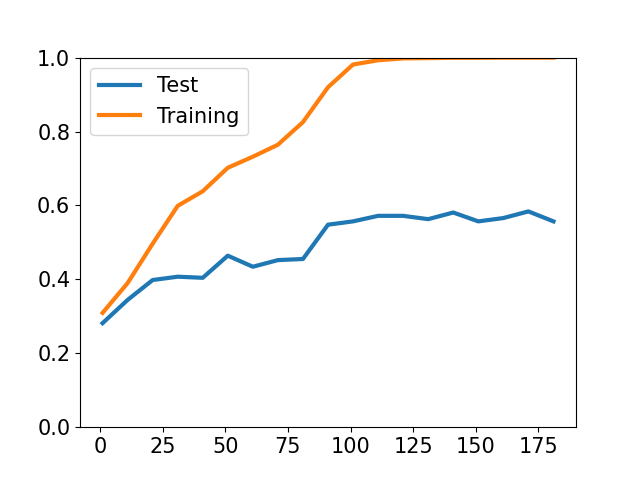}%
\hfill
\includegraphics[width = 0.5\textwidth]{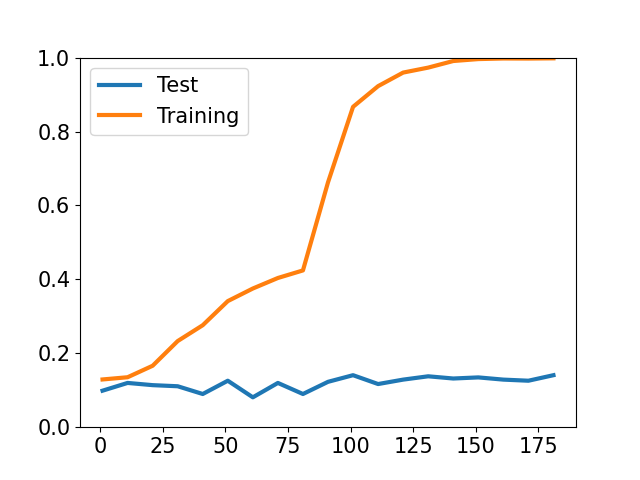}
\end{minipage}
}
\caption{Clean Accuracy and Adversarial Accuracy on \textbf{Atypical} Set of CIFAR10}
\label{fig:app_1_22}
\vspace{-0.5cm}
\end{figure}

\begin{figure}[h!]
\centering
\hspace*{-1cm}
\subfloat[Clean (left) \& Adv Acc. (right) under ResNet32.]{
\begin{minipage}[h]{0.55\textwidth}
\includegraphics[width = 0.5\textwidth]{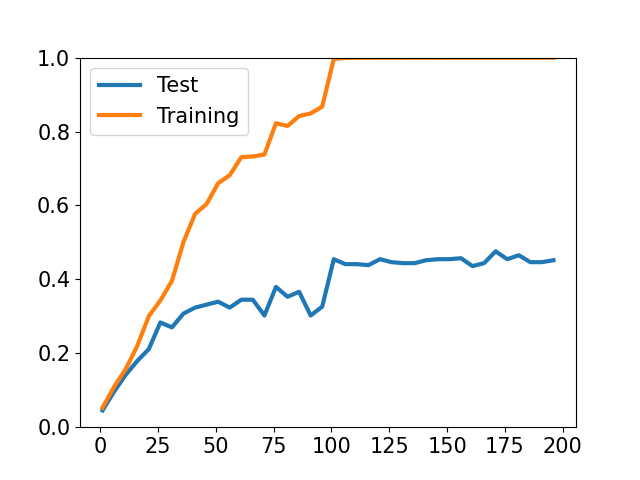}%
\hfill
\includegraphics[width = 0.5\textwidth]{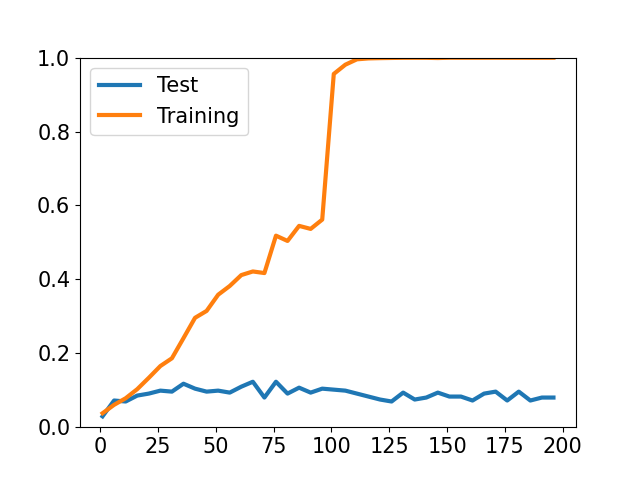}
\end{minipage}
}
\hspace*{-0.4cm}
\subfloat[Clean (left) \& Adv Acc. (right) under WRN28.]{
\begin{minipage}[c]{0.55\textwidth}
\includegraphics[width = 0.5\textwidth]{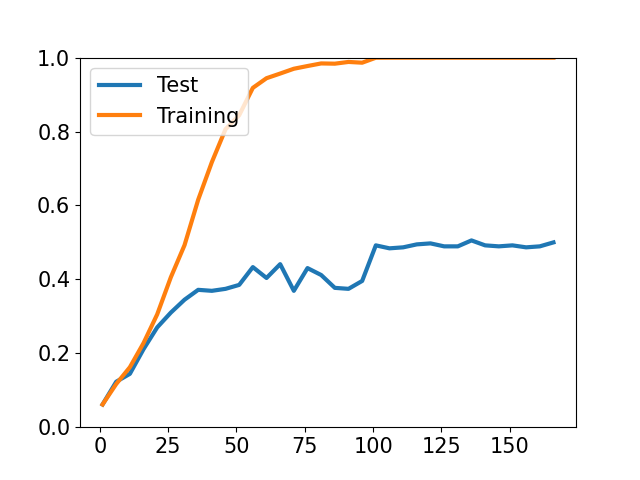}%
\hfill
\includegraphics[width = 0.5\textwidth]{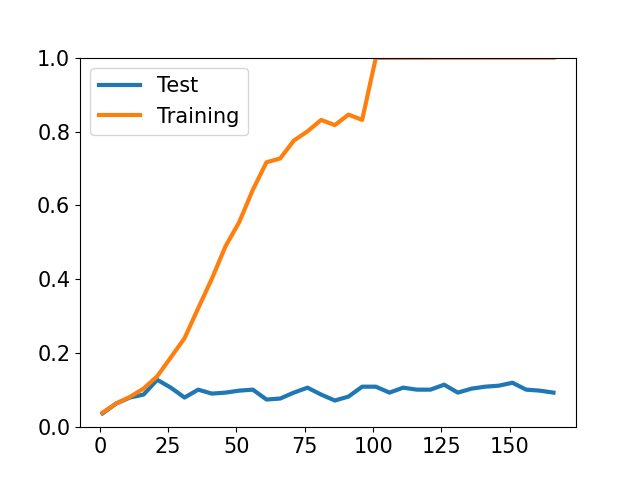}
\end{minipage}
}
\caption{Clean Accuracy and Adversarial Accuracy on \textbf{Atypical} Set of Tiny~ImageNet}
\label{fig:app_1_23}
\end{figure}

Fig.~\ref{fig:app_1_21}, Fig.~\ref{fig:app_1_22} and Fig.~\ref{fig:app_1_23} report the performance of adversarially trained models. We evaluate the clean accuracy and adversarial accuracy on the training atypical set $\mathcal{D}_\text{atyp}$ and test atypical set $\mathcal{D}'_\text{atyp}$. From the results, we can observe that although fitting atypical samples can help the model to have modest clean accuracy on test atypical samples, the adversarial robustness of them is constantly low and can hardly be improved during the whole training process.

\subsection{Additional Results for Preliminary Study - Section~\ref{sec:pre2}}\label{app:pre2}

In this subsection, we provide more experimental results to validate the statement 
in Section~\ref{sec:pre2}, where we state that fitting atypical samples in adversarial training can hurt the performance (clean \& adversarial accuracy) of typical samples.
We provide full empirical results to show that: \textbf{(i)} In traditional ERM, fitting atypical samples will not hurt the models' performance (clean accuracy) of typical samples. \textbf{(ii)} In adversarial training, fitting atypical samples can degrade the clean \& adversarial accuracy of typical samples. \textbf{(iii)} In adversarial training, fitting atypical samples can degrade the quality of learned representations, especially the models' discrimination between classes.
The experimental setting follows Section~\ref{sec:pre2}, where we train the models for several trails on resampled (CIFAR100, CIFAR10, Tiny~ImageNet) datasets: each dataset is constructed with the whole training typical set $\mathcal{D}_\text{typ}$, and a part of the training atypical set $\mathcal{D}_\text{atyp}$ (randomly sample 0\%, 20\% and 100\% in $\mathcal{D}_\text{atyp}$). We evaluate the models' performance on test ``typical'' set $\mathcal{D}'_\text{typ}$ which is defined in Section~\ref{sec:pre2}.

\newpage
\textbf{(i) Additional Results for Preliminary Study - Section~\ref{sec:pre2} In Traditional ERM}

\begin{figure}[h]
\centering
\hspace*{-1cm}
\subfloat[ResNet18.]{
\begin{minipage}[c]{0.3\textwidth}
\includegraphics[width = 1.0\textwidth]{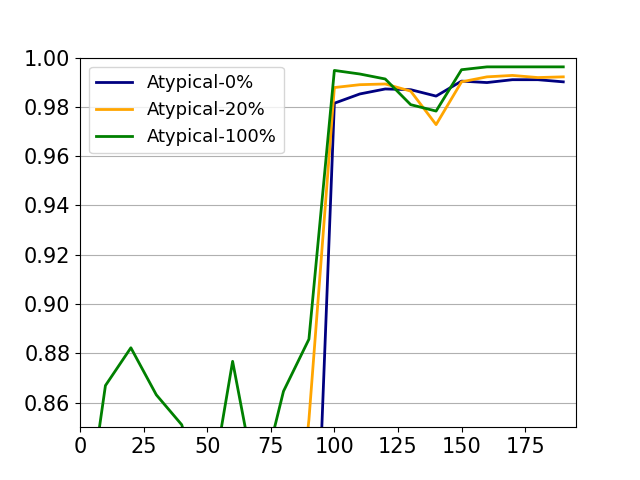}
\end{minipage}
}
\hspace*{0.4cm}
\subfloat[WRN28.]{
\begin{minipage}[c]{0.3\textwidth}
\includegraphics[width = 1.0\textwidth]{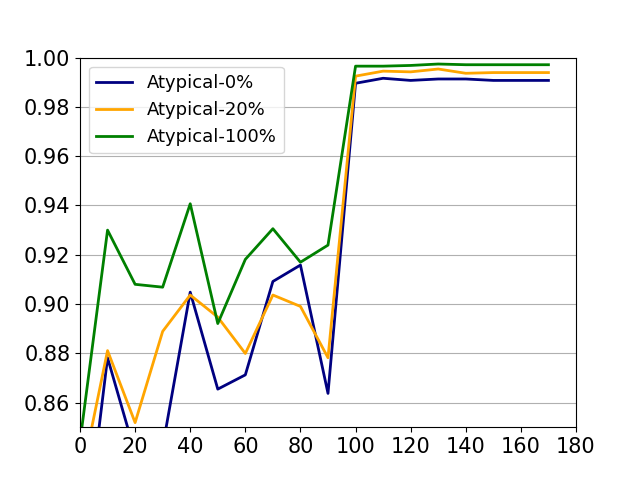}%
\end{minipage}
}
\caption{Clean Accuracy on \textbf{Typical} Set of CIFAR100}
\label{fig:app2_11}
\vspace{-0.5cm}
\end{figure}

\begin{figure}[h]
\centering
\hspace*{-1cm}
\subfloat[ResNet18.]{
\begin{minipage}[c]{0.3\textwidth}
\includegraphics[width = 1.0\textwidth]{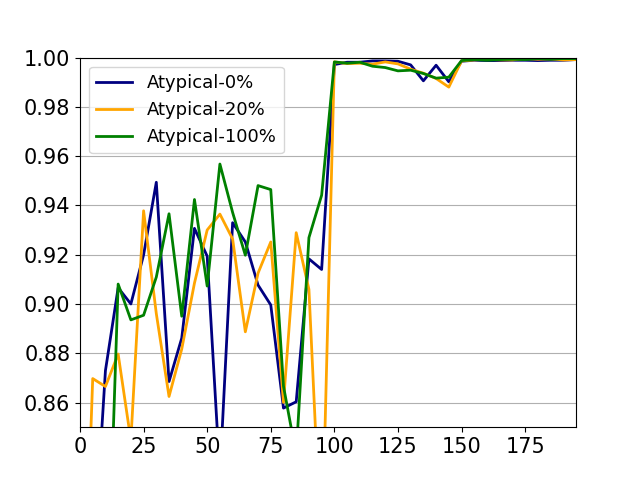}
\end{minipage}
}
\hspace*{0.4cm}
\subfloat[WRN28.]{
\begin{minipage}[c]{0.3\textwidth}
\includegraphics[width = 1.0\textwidth]{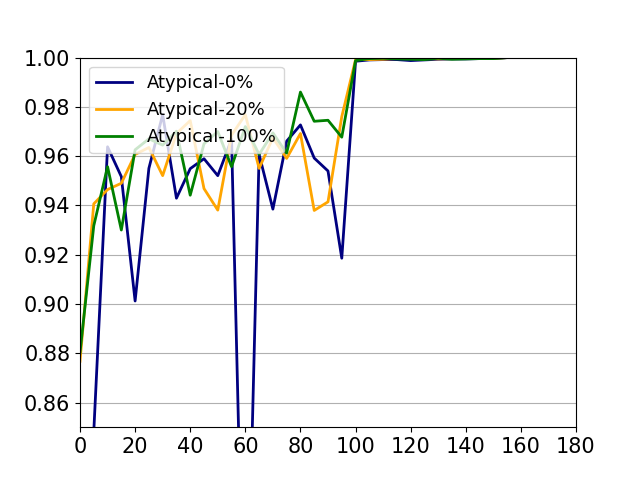}%
\end{minipage}
}
\caption{Clean Accuracy on \textbf{Typical} Set of CIFAR10}
\label{fig:app2_12}
\vspace{-0.5cm}
\end{figure}

\begin{figure}[h!]
\centering
\hspace*{-1cm}
\subfloat[ResNet32.]{
\begin{minipage}[c]{0.3\textwidth}
\includegraphics[width = 1.0\textwidth]{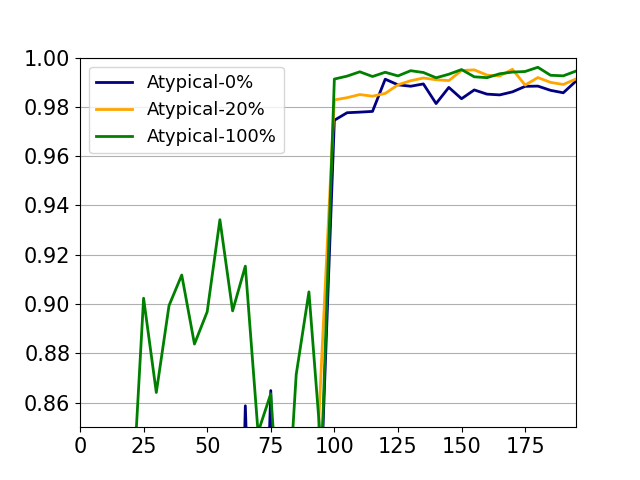}
\end{minipage}
}
\hspace*{0.4cm}
\subfloat[WRN28.]{
\begin{minipage}[c]{0.3\textwidth}
\includegraphics[width = 1.0\textwidth]{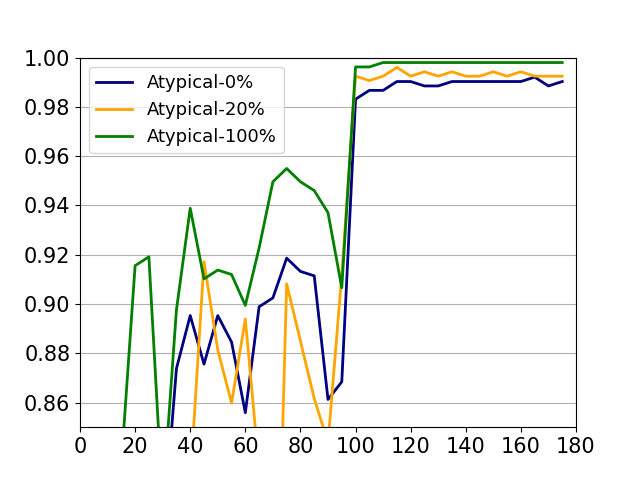}%
\end{minipage}
}
\caption{Clean Accuracy on \textbf{Typical} Set of Tiny~ImageNet}
\label{fig:app2_13}
\end{figure}

Fig.~\ref{fig:app2_11}, Fig.~\ref{fig:app2_12} and Fig.~\ref{fig:app2_13} report the performance of traditional ERM, trained on different resampled datasets with different amount of atypical samples existed. The figures report the clean accuracy on test typical set of CIFAR100, CIFAR10 and Tiny~ImageNet. We also leave the robustness performance out here as the models are not robust to adversarial attacks. From the results, we can conclude that in traditional ERM, fitting atypical samples will not degrade the models' performance on typical samples. For example, in CIFAR100 dataset, with 100\% atypical samples included (Atypical-100\%)., the accuracy on the test typical set is even slightly higher than the model trained without atypical samples (Atypical-0\%). This conclusion is consistent for all three datasets and model architectures.


\textbf{(ii) Additional Results for Preliminary Study - Section~\ref{sec:pre2} In Adversarial Training}

\begin{figure}[h]
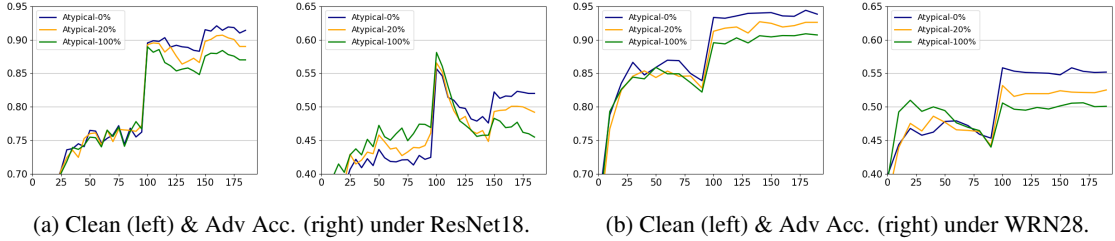
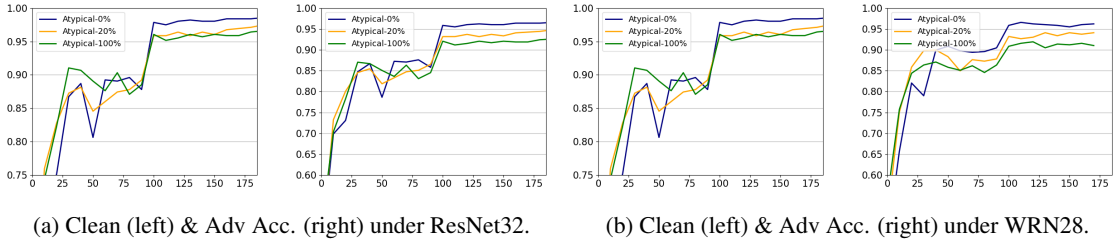

\centering
\hspace*{-1cm}
\subfloat[Clean (left) \& Adv Acc. (right) under ResNet18.]{
\begin{minipage}[h]{0.55\textwidth}
\includegraphics[width = 0.5\textwidth]{figures/poison_clean_ResNet18.png}%
\hfill
\includegraphics[width = 0.5\textwidth]{figures/poison_adv_ResNet18.png}
\end{minipage}
}
\hspace*{-0.4cm}
\subfloat[Clean (left) \& Adv Acc. (right) under WRN28.]{
\begin{minipage}[c]{0.55\textwidth}
\includegraphics[width = 0.5\textwidth]{figures/poison_clean_WRN.png}%
\hfill
\includegraphics[width = 0.5\textwidth]{figures/poison_adv_WRN.png}
\end{minipage}
}
\caption{Clean Accuracy and Adversarial Accuracy on \textbf{Typical} Set of CIFAR100}
\label{fig:pre2_21}
\vspace{-0.3cm}
\end{figure}

\begin{figure}[h]
\centering
\hspace*{-1cm}
\subfloat[Clean (left) \& Adv Acc. (right) under ResNet18.]{
\begin{minipage}[h]{0.55\textwidth}
\includegraphics[width = 0.5\textwidth]{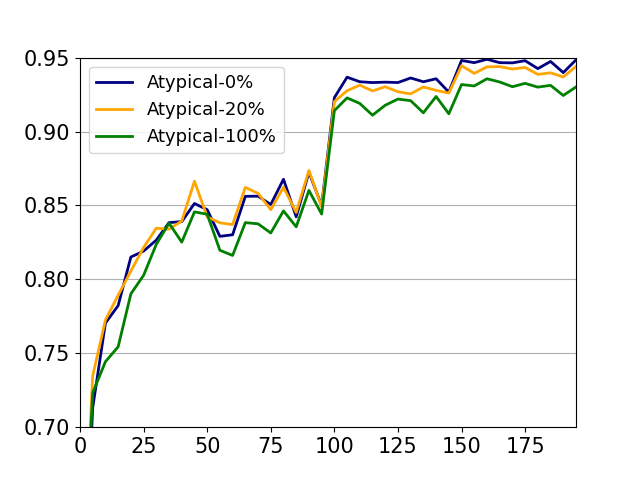}%
\hfill
\includegraphics[width = 0.5\textwidth]{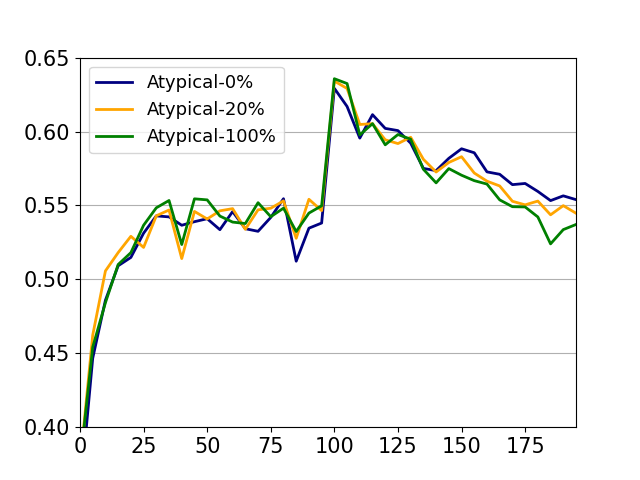}
\end{minipage}
}
\hspace*{-0.4cm}
\subfloat[Clean (left) \& Adv Acc. (right) under WRN28.]{
\begin{minipage}[c]{0.55\textwidth}
\includegraphics[width = 0.5\textwidth]{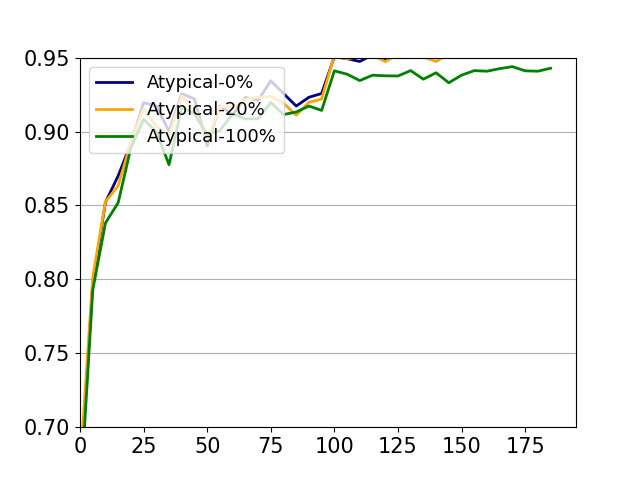}%
\hfill
\includegraphics[width = 0.5\textwidth]{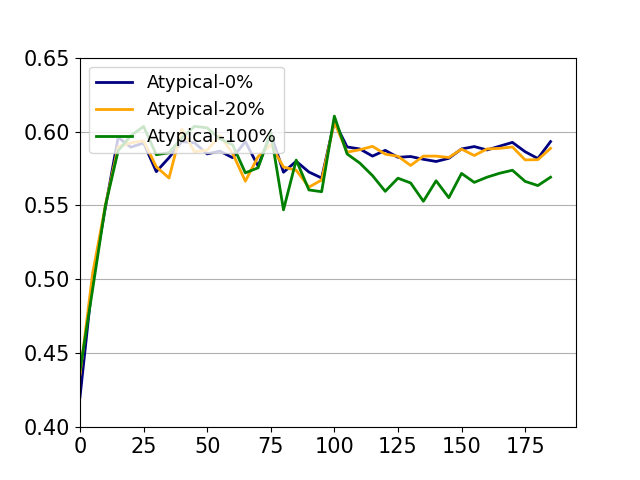}
\end{minipage}
}
\caption{Clean Accuracy and Adversarial Accuracy on \textbf{Typical} Set of CIFAR10}
\label{fig:pre2_22}
\vspace{-0.3cm}
\end{figure}

\begin{figure}[h!]
\centering
\hspace*{-1cm}
\subfloat[Clean (left) \& Adv Acc. (right) under ResNet32.]{
\begin{minipage}[h]{0.55\textwidth}
\includegraphics[width = 0.5\textwidth]{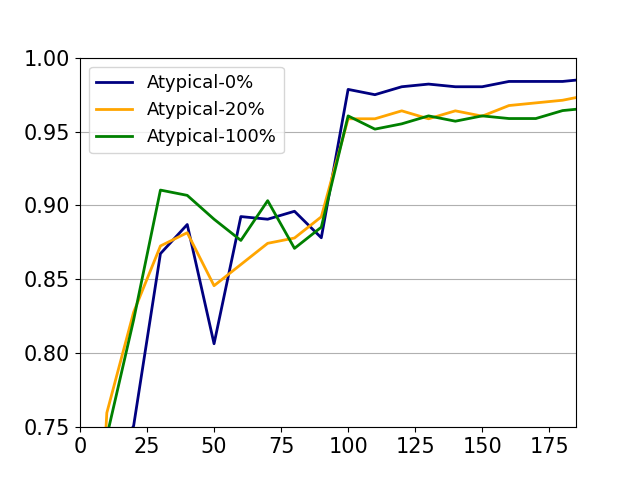}%
\hfill
\includegraphics[width = 0.5\textwidth]{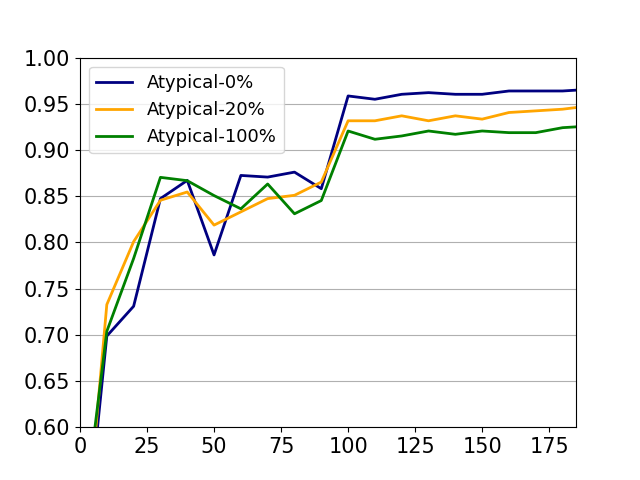}
\end{minipage}
}
\hspace*{-0.4cm}
\subfloat[Clean (left) \& Adv Acc. (right) under WRN28.]{
\begin{minipage}[c]{0.55\textwidth}
\includegraphics[width = 0.5\textwidth]{figures/pre3_imagenet_adv1_resnet18.png}%
\hfill
\includegraphics[width = 0.5\textwidth]{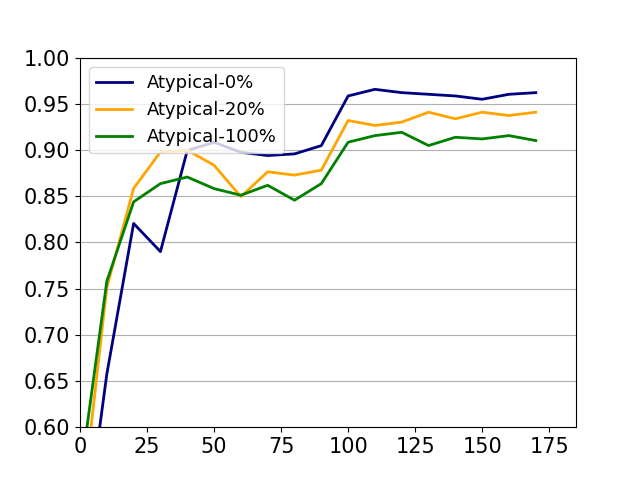}
\end{minipage}
}
\caption{Clean Accuracy and Adversarial Accuracy on \textbf{Typical} Set of CIFAR100}
\label{fig:pre2_23}
\end{figure}

Fig.~\ref{fig:pre2_21}, Fig.~\ref{fig:pre2_22} and Fig.~\ref{fig:pre2_23} report the performance of adversarial training, on different resampled datasets with different amount of atypical samples existed. The figures report both clean and adversarial accuracy on test atypical sets of CIFAR100, CIFAR10 and Tiny~ImageNet. Based on the experimental results, we find that including more atypical samples can cause the model have worse performance on typical samples in all three datasets. In datasets with a large portion of atypical samples, such as CIFAR100, the negative effects of atypical samples are more obvious. In CIFAR100, training on datasets with 100\% atypical samples (Atypical 100\%) can cause the clean \& adversarial accuracy drop by $\sim7\%$ and $8\%$, respectively.


\vspace{2cm}
\textbf{(iii) Additional Results for Preliminary Study - Section~\ref{sec:pre2} Classwise Representation Distance}

\begin{figure}[h]
\centering
\hspace*{-1cm}
\subfloat[Traditional ERM under ResNet18 (left), WRN28 (right).]{
\begin{minipage}[h]{0.55\textwidth}
\includegraphics[width = 0.5\textwidth]{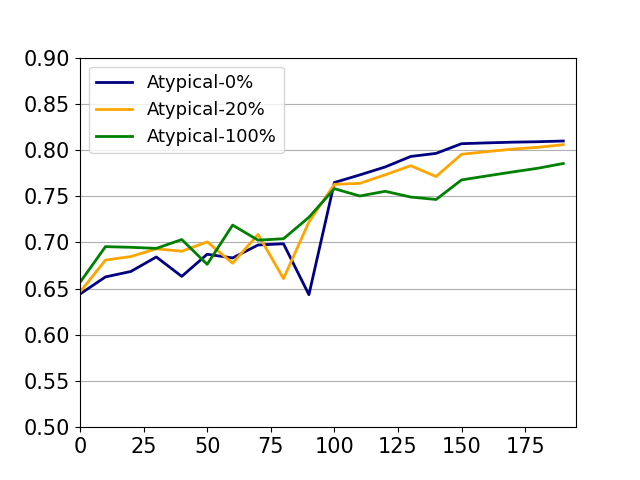}%
\hfill
\includegraphics[width = 0.5\textwidth]{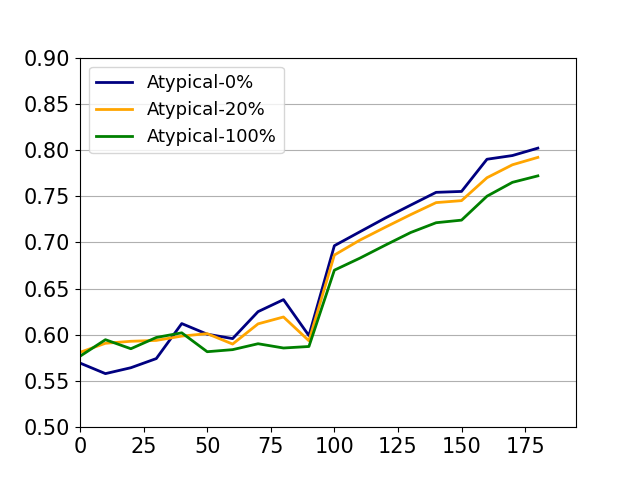}
\end{minipage}
}
\hspace*{-0.4cm}
\subfloat[Adv. Training under ResNet18 (left), WRN28 (right).]{
\begin{minipage}[c]{0.55\textwidth}
\includegraphics[width = 0.5\textwidth]{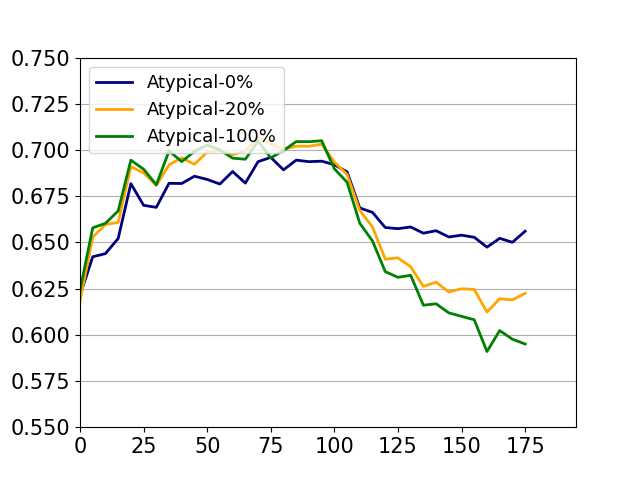}%
\hfill
\includegraphics[width = 0.5\textwidth]{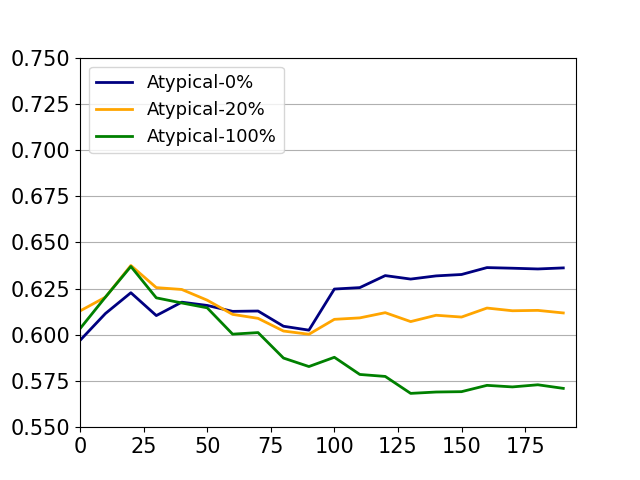}
\end{minipage}
}
\caption{Classwise Cosine Distance of Output Representation of \textbf{Typical} Set of CIFAR100}
\label{fig:dist1}
\end{figure}

\begin{figure}[h]
\centering
\hspace*{-1cm}
\subfloat[Traditional ERM under ResNet18 (left), WRN28 (right).]{
\begin{minipage}[h]{0.55\textwidth}
\includegraphics[width = 0.5\textwidth]{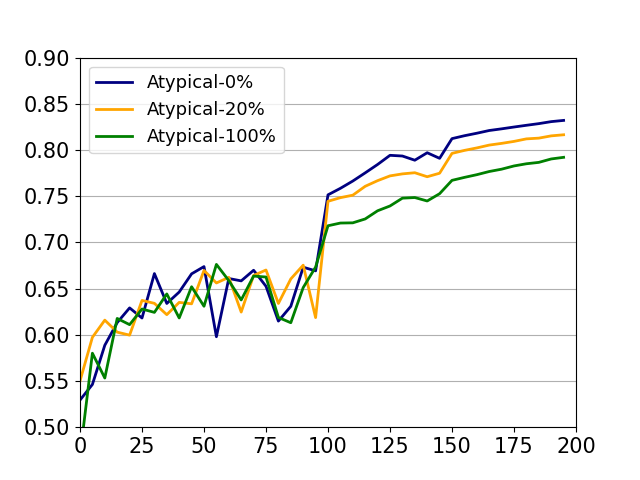}%
\hfill
\includegraphics[width = 0.5\textwidth]{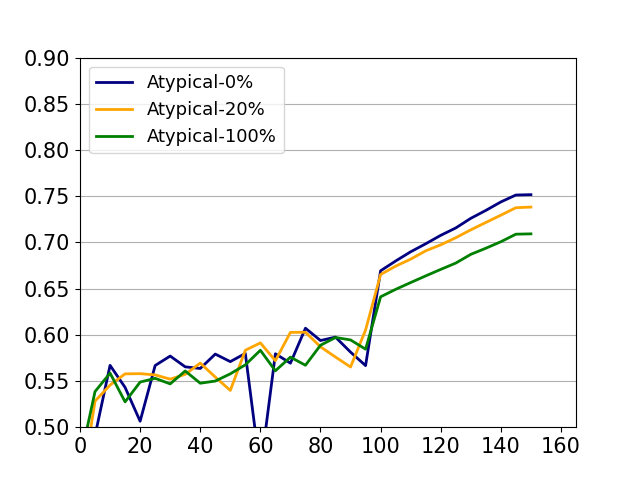}
\end{minipage}
}
\hspace*{-0.4cm}
\subfloat[Adv. Training under ResNet18 (left), WRN28 (right).]{
\begin{minipage}[c]{0.55\textwidth}
\includegraphics[width = 0.5\textwidth]{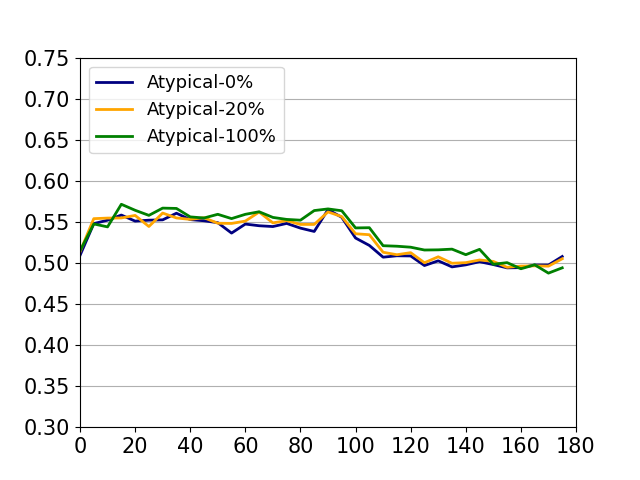}%
\hfill
\includegraphics[width = 0.5\textwidth]{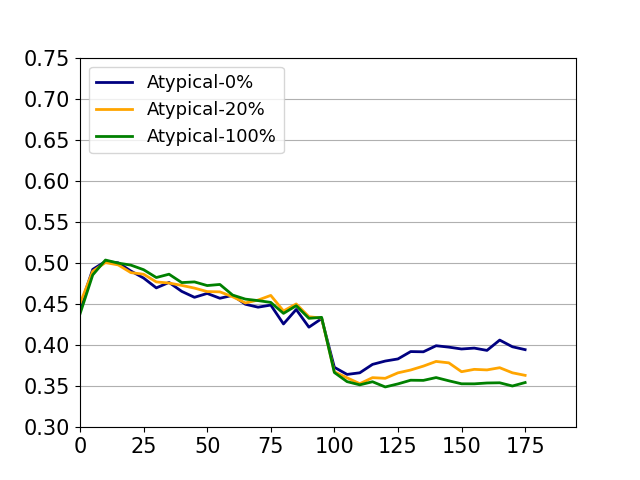}
\end{minipage}
}
\caption{Classwise Cosine Distance of Output Representation of \textbf{Typical} Set of CIFAR10}
\label{fig:dist2}
\end{figure}

\begin{figure}[h!]
\centering
\hspace*{-1cm}
\subfloat[Traditional ERM under ResNet32 (left), WRN28 (right).]{
\begin{minipage}[h]{0.55\textwidth}
\includegraphics[width = 0.5\textwidth]{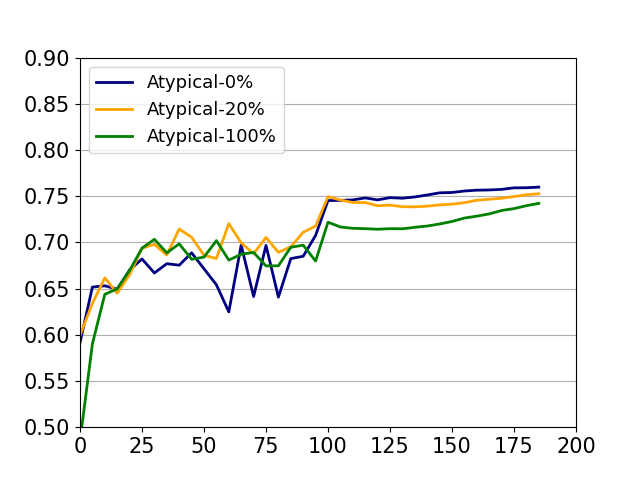}%
\hfill
\includegraphics[width = 0.5\textwidth]{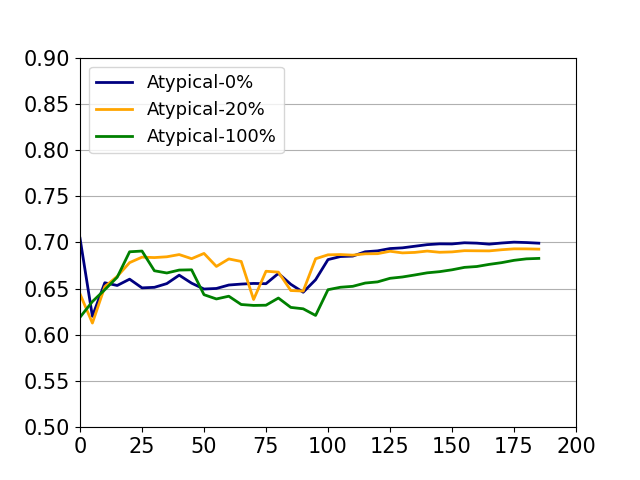}
\end{minipage}
}
\hspace*{-0.4cm}
\subfloat[Adv. Training under ResNet32 (left), WRN28 (right).]{
\begin{minipage}[c]{0.55\textwidth}
\includegraphics[width = 0.5\textwidth]{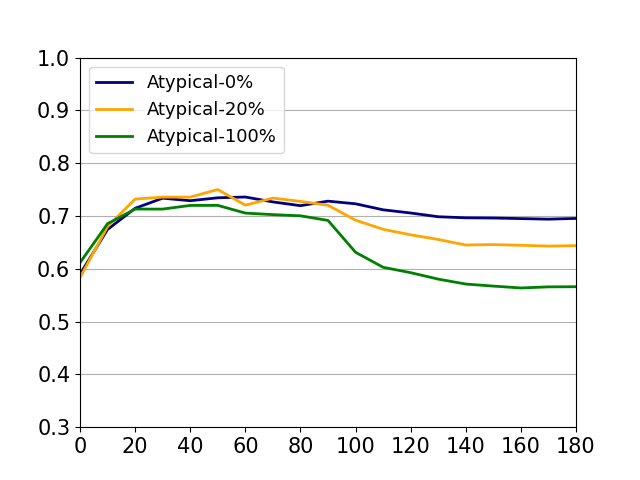}%
\hfill
\includegraphics[width = 0.5\textwidth]{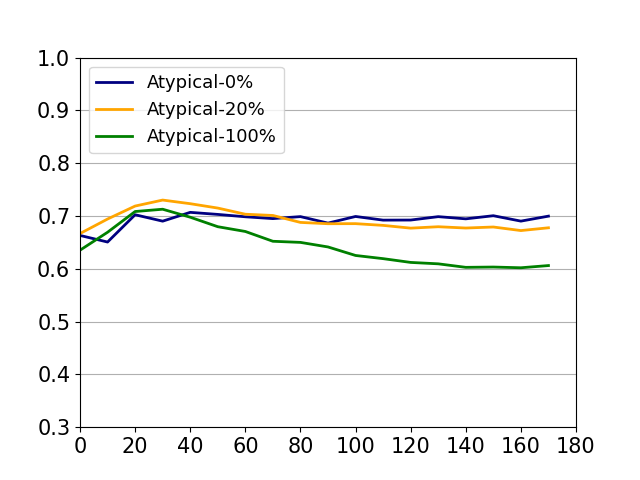}
\end{minipage}
}
\caption{Classwise Cosine Distance of Output Representation of \textbf{Typical} Set of Tiny~ImageNet}
\label{fig:dist3}
\end{figure}

Here, we provide additional evidence to show that in adversarial training, fitting atypical samples can degrade the quality of DNN's learned representations (as proposed in Section~\ref{sec:pre2}). In Fig.~\ref{fig:dist1}, Fig.~\ref{fig:dist2} and Fig.~\ref{fig:dist3}, we measure the Cosine Distance (defined in Section~\ref{sec:pre2}) of the representations for (typical) samples from different classes. In these figures, we provide detailed results of both traditional ERM and adversarial training under ResNet18 and WRN28. From the results, we can see that in adversarial training, more atypical samples can cause the smaller classwise Cosine Distance of the models (on their last epochs). Moreover, during the training process of adversarial training, the classwise Cosine Distance first increases and then starts to decrease, especially when there are more atypical samples are fitted. For example, in CIFAR100 under ResNet18, the Cosine Distance starts to decrease at around Epoch 100, which is when many more atypical samples are fitted. These results suggest that, in adversarial training, fitting atypical samples can be an important reason to cause the models to produce poor representations. As a comparison, in Traditional ERM, the classwise Cosine Distance keeps increasing from the first epoch to the last one. Although given the same epoch, the models trained with more atypical samples also have smaller Cosine Distance, they would not harm the model's final performance. It is because the models trained with more atypical samples can also have relatively large classwise Cosine Distance in their last epochs.

\section{The Detailed Training Scheme of BAT}\label{app:algorithm}

In this section, we provide the detailed training scheme of BAT in Algorithm~\ref{alg:bat}. In particular, BAT algorithm starts from a randomly initialized neural network. On each mini-batch, it applies PGD attack to generate (training) adversarial examples (Step 5). Following the Eq~\ref{eq:poisoning_score} and Eq~\ref{eq:reweight}, BAT calculates which samples are likely to be \textit{Poisoning Atypical Samples} and their corresponding weight values (Step 6). Under the current mini-batch, next, BAT calculates the \textit{Discrimination Loss} of the typical samples $\mathcal{D}_\text{typ}$ (Step 8). Finally, BAT uses SGD to update the model parameter to minimize the reweighted adversarial loss regularized by $\beta$ times discrimination loss (Step 8).

\begin{algorithm}[h!]
\begin{algorithmic}[1]
\setstretch{1}
\STATE \textbf{Input:} Training dataset $\mathcal{D}$, with typical set $\mathcal{D}_\text{typ} = \{x \in\mathcal{D}; \text{mem}(x) \leq \sigma\}$, atypical set $\mathcal{D}_\text{atyp} = \{x \in\mathcal{D}; \text{mem}(x) > \sigma\}$. Targeted type of adversarial attack: $l_\infty$-$\epsilon$ attack.
Hyperparameters $\alpha, \beta \in \R^+$. \\
\STATE Randomly initialize the network $F$ \\
\REPEAT
\STATE Fetch mini-batch data $\{(x_i,y_i)\}$ at current epoch\\
\STATE Using PGD to generate adversarial training sample  $\{(x_i^\text{adv},y_i)\}$\\
\STATE Calculate Poisoning Score $\textbf{q}(x^\text{adv}_i)$ and weight $w_i$ as in Equation~\ref{eq:poisoning_score} and Equation~\ref{eq:reweight}.
\STATE Calculate Discrimination Loss $\mathcal{L}_{DL}(F)$ within the current mini-batch, as in Equation~\ref{eq:dl}.
\STATE Update $F$ by SGD on the objective: $\mathcal{L}_\text{BAT} = \frac{1}{\sum_i w_i }\sum_i \left[w_i \cdot \mathcal{L}(F(x_i^\text{adv}), y_i)\right] + \beta\cdot\mathcal{L}_{DL}(F)$.
\UNTIL{End of Training}
\caption{The Benign Adversarial Training (BAT) Algorithm}
\label{alg:bat}
\end{algorithmic}
\end{algorithm}

\section{Additional Results for Experiment}\label{app:exp}

In this section, we provide additional experimental results to validate the effectiveness of BAT method. In Table~\ref{tab:resnet18_cifar100} and Table~\ref{tab:wrn28_cifar100}, we provide the results of BAT and baseline models on CIFAR100 dataset under ResNet18 and WRN28 architectures. In Table~\ref{tab:resnet18_imagenet} and Table~\ref{tab:wrn28_imagenet}, we provide the results of BAT and baseline models on Tiny~ImageNet dataset under ResNet32 and WRN28 architectures. In the experiments, we train the models for 160 epochs with learning rate 0.1, momentum 0.9, weight decay 5e-4, and decay the learning rate by 0.1 at the epoch 80 and 120. To have a more comprehensive and reliable adversarial robustness, in addition to PGD adversarial attack~\cite{madry2017towards}, we also measure the model's adversarial accuracy via other attack algorithms, including FGSM attack~\cite{goodfellow2014explaining}, CW attack~\cite{carlini2017towards} and Auto Attack~\cite{croce2020reliable}. The results show that the BAT method can consistently outperform baseline models, as BAT has better clean accuracy vs. adversarial accuracy trade-off. Under ResNet18, the BAT method achieves comparable adversarial accuracy to the best baseline methods, and BATs have the highest clean accuracy. Under WRN28, the BAT method has both higher clean \& adversarial accuracy than the baseline methods.

\begin{table}[h]
\small
\centering
\caption{Performance of BAT vs. Baselines on CIFAR100 Under ResNet18}
\label{tab:resnet18_cifar100}
\begin{tabular}{c|c|ccccc}
\hline
Method & Clean Acc. & FGSM & PGD & CW & AA. \\
\hline
\hline
PGD Train (Best Adv.) & 56.9 & 36.0 & 27.4 & 25.4 & 23.6 \\
PGD Train (Best Clean) & 57.8 & 33.5 & 21.9 & 22.5 & 20.2 \\
TRADES ($1/\lambda = 5$) & 56.6 & 36.5 & 26.9 & 25.3 & 23.9 \\
MART~\cite{wang2019improving} & 51.8 & 36.1 & \textbf{30.4} & 25.8 & \textbf{24.4} \\
GAIRAT~\cite{zhang2020geometry} & 58.2 &36.5 & 27.8 & 25.9 & 23.8\\
\hline
BAT ($\alpha = 1, \beta = 0.2$) & \textbf{59.5} & \textbf{37.3} & 27.3 & \textbf{26.6} & 24.3\\
BAT ($\alpha = 2, \beta = 0.2$) &59.3 & 37.1 & 27.4 & 26.5 & 24.0\\
\hline
\hline
\end{tabular}
\end{table}

\begin{table}[h]
\small
\centering
\caption{Performance of BAT vs. Baselines on CIFAR100 Under WRN28}
\label{tab:wrn28_cifar100}
\begin{tabular}{c|c|ccccc}
\hline
Method & Clean Acc. & FGSM & PGD & CW & AA. \\
\hline
\hline
PGD Train (Best Adv.) & 59.7 & 34.9 & 24.7 & 24.2 & 22.5 \\
PGD Train (Best Clean.) & 59.7 & 34.9 & 24.7 & 24.2 & 22.5 \\
TRADES ($1/\lambda = 5$) & 57.3 & 34.5 & 24.9 & 24.6 & 22.9 \\
MART~\cite{wang2019improving} & 56.5 & 36.1 & 26.8 & 25.3 & 23.8 \\
GAIRAT~\cite{zhang2020geometry} & 60.2 & 34.8 & 24.4 & 24.8 & 22.9 \\
\hline
BAT ($\alpha = 1, \beta = 0.2$) & \textbf{62.0} & 38.6 & \textbf{28.5} & \textbf{26.5} & \textbf{24.8} \\
BAT ($\alpha = 2, \beta = 0.2$) & 61.4 & \textbf{38.9} & 28.2 & 26.3 & \textbf{24.8} \\
\hline
\hline
\end{tabular}
\end{table}

\begin{table}[h]
\small
\centering
\caption{Performance of BAT vs. Baselines on Tiny~ImageNet Under ResNet32}
\label{tab:resnet18_imagenet}
\begin{tabular}{c|c|ccccc}
\hline
Method & Clean Acc. & FGSM & PGD & CW & AA. \\
\hline
\hline
Adv. Train (Best Adv.) & 56.3 & 37.5 & 32.3 & 29.8 & 29.8\\
Adv. Train (Best Clean) & 58.2 & 36.8 & 30.5 & 28.8 & 28.4\\
TRADES ($1/\lambda = 5$) & 55.4 & 35.2 & 28.8 & 27.0 & 27.0\\
MART~\cite{wang2019improving} & 56.2 & 38.1& \textbf{34.5} & \textbf{31.8} & 32.0\\
GAIRAT~\cite{zhang2020geometry} & 58.4 & 37.3& 30.4 & 28.9 & 29.0\\
\hline
BAT ($\alpha = 1, \beta = 0.2$) & \textbf{59.4} &40.4 &32.0 & 31.5 & 32.0\\
BAT ($\alpha = 2, \beta = 0.2$) & \textbf{59.4} & \textbf{41.3}  &32.9 & \textbf{31.8} & \textbf{32.4}\\
\hline
\hline
\end{tabular}
\end{table}
\begin{table}[h!]
\small
\centering
\caption{Performance of BAT vs. Baselines on Tiny~ImageNet Under WRN28}
\label{tab:wrn28_imagenet}
\begin{tabular}{c|c|ccccc}
\hline
Method & Clean Acc. & FGSM & PGD & CW & AA. \\
\hline
\hline
Adv. Train (Best Adv.) & 58.9 & 35.7 & 31.7 & 30.0 & 30.0  \\
Adv. Train (Best Clean) & 60.0 & 35.1 & 30.1 & 28.8 & 28.5\\
TRADES ($1/\lambda = 5$) & 59.7 & 37.4 & 32.0 & 31.8 & 32.0\\
MART~\cite{wang2019improving} & 58.2 & 41.0 & 35.6 & 33.9 & 34.1 \\
GAIRAT~\cite{zhang2020geometry} & 59.9 & 38.3 & 32.4 & 31.0 & 31.1\\
\hline
BAT ($\alpha = 1, \beta = 0.2$) & 62.3 & 41.6 & 35.8 & 33.7 & 34.3 \\
BAT ($\alpha = 2, \beta = 0.2$) & \textbf{62.4} & \textbf{43.1} &\textbf{37.4} & \textbf{35.3} & \textbf{35.6}\\
\hline
\hline
\end{tabular}
\end{table}

\section{Boarder Impact}\label{app:board}

Nowadays, deep neural networks (DNNs) have been widely applied to solve various machine learning tasks, especially on many safety-critical tasks such as autonomous vehicle~\cite{fagnant2015preparing}, AI healthcare~\cite{jiang2017artificial} and ID authentication~\cite{mohammed2011human}, etc. However, the existence of adversarial attacks~\cite{xu2019adversarial} brings huge threats to the safety of these DNNs' applications. As one of the most successful approaches to defend DNNs against adversarial attacks, adversarial training methods~\cite{madry2017towards, zhang2016understanding} still suffer from several disadvantages. For example, for a DNN model to achieve good adversarial robustness, it usually sacrifices its clean accuracy. Adversarially trained DNNs also present strong overfitting effects. In our study, we draw important findings to explain these drawbacks of adversarial training from the data perspective and empirically validate these issues' relation to atypical samples in the data distribution. Moreover, the currently existed adversarially trained models can only achieve satisfactory performance on simple datasets such as MNIST and CIFAR10. Motivated from our findings, we propose a new method to improve the performance of adversarial training, especially on more complex datasets. We anticipate that our findings and method are helpful for further studies to improve the effectiveness and feasibility of adversarial training and eventually build safer DNNs.

\end{document}